\documentclass{article}

\PassOptionsToPackage{numbers,sort&compress}{natbib}

\usepackage[utf8]{inputenc}
\usepackage[T1]{fontenc}

\usepackage{microtype}
\usepackage{graphicx}
\usepackage{booktabs} 
\usepackage{multirow}
\usepackage{threeparttable}
\usepackage[table]{xcolor}
\usepackage{subcaption}
\usepackage{colortbl}     
\usepackage{makecell}
\usepackage{wrapfig}
\usepackage{tikz}
\usepackage{pifont}
\usepackage{amsthm}
\usepackage{algorithm}
\usepackage{algpseudocode}
\usepackage{bm}
\definecolor{bgblue}{RGB}{200,225,240}   
\definecolor{bgred}{RGB}{245,200,200}    
\definecolor{sectionblue}{RGB}{220,235,242}

\definecolor{AuthorTwoColor}{RGB}{204, 0, 102}   

\newcommand{\teasercell}[1]{\includegraphics[width=\dimexpr\linewidth-2\fboxrule\relax]{#1}}
\newcommand{\teaserHdr}[1]{{\fontsize{7}{8}\selectfont\color[gray]{0.25}\textbf{#1}}}
\newcommand{\teaserDoLabel}[1]{{\fontsize{8}{10}\selectfont #1}}
\newcommand{\teaserSubcap}[1]{{\fontsize{8}{9.5}\selectfont\textbf{#1}}}
\usepackage{hyperref}

\providecommand*{\rvx}{\mathbf{x}}
\newcommand*{\x}{\rvx}
\newcommand*{\z}{\mathbf{z}}
\newcommand*{\cc}{\mathbf{c}}

\newcommand*{\pa}{\mathbf{pa}}

\newcommand*{\pacf}{\widetilde{\pa}}
\newcommand*{\doo}{\operatorname{do}}
\newcommand{\cmt}[1]{}

\newcommand*{\eps}{\boldsymbol{\epsilon}}

\newcommand*{\uu}{\mathbf{u}}

\newcommand*{\indep}{\perp \!\!\! \perp}

\newcommand{\identity}{\mathbf{I}}

\newcommand{\spec}[1]{\textnormal{\textsc{#1}}}

\usepackage[preprint]{neurips_2026}

\usepackage{amsmath}
\usepackage{amssymb}
\usepackage{mathtools}

\usepackage[capitalize,noabbrev]{cleveref}

\theoremstyle{plain}

\theoremstyle{definition}

\theoremstyle{remark}

\usepackage[disable,textsize=tiny]{todonotes}

\makeatletter
\newcommand\blfootnote[1]{%
  \begingroup
  \renewcommand\thefootnote{}%
  \begin{NoHyper}\footnote{#1}\end{NoHyper}%
  \addtocounter{footnote}{-1}%
  \endgroup
}
\makeatother

\begin{document}

\title{Generating Chest X-Ray Counterfactuals by Specialising Foundation Image Models}

%

\author{%
  Xiaodan Xing \textsuperscript{1,*} \qquad
  Rajat R.~Rasal \textsuperscript{2, *,\dag} \qquad
  Julia A.~Meister \textsuperscript{1, *} \\
  \bf
  Sara Ghorayeb \textsuperscript{1} \qquad
  Galvin Khara \textsuperscript{1} \qquad
  Jessica Schrouff \textsuperscript{1} \\[3pt]
  \textsuperscript{1}GSK \qquad \textsuperscript{2}Imperial College London \\[2pt]
  \texttt{xiaodan.x.xing@gsk.com}
}

\maketitle
\blfootnote{\textsuperscript{*}Equal contribution.}
\blfootnote{\textsuperscript{\dag}Work performed during an internship at GSK.}
\begin{abstract}
Counterfactual image generation answers questions about how a subject would have looked under retrospective, hypothetical scenarios.
Recent methods have improved perceptual quality, identity preservation and faithfulness to an underlying causal model, but their adoption in healthcare is limited by scarce annotated data, distribution shift between datasets, and mismatches between pretrained generative models and those required for counterfactual inference.
We propose \textbf{\emph{specialisation}}, a data and parameter-efficient framework for adapting pretrained, non-causal generative models into causal mechanisms under distribution shift.
Based on this framework, we train a radiology counterfactual image generation model, called RadCF, using latent flow matching.
We validate our approach on three chest X-ray datasets spanning different dataset shifts, data volumes, and counterfactual questions, associated with challenging, highly-localised interventions.
Our results show that RadCF and specialisation improve counterfactual soundness over existing methods while being data and parameter efficient, and that the resulting counterfactuals can detect and mitigate shortcut learning in a downstream medical classifier.
Code is available at \url{https://github.com/GSK-AI/RadCF/}.
\end{abstract}

\section{Introduction}
\label{sec:introduction}

Advances in generative modelling have improved faithfulness and control in medical image synthesis, particularly for chest X-rays \citep{Bluethgen2024, perez-garcia_bond-taylor_radedit, weber2023cascaded, pinaya2023generative, trang2025discovering}.
However, these models often struggle with fine-grained, \emph{causal} edits, as they implicitly learn broad statistical associations between dense conditioning signals and images.
In the healthcare domain, models require an understanding of how genetic, environmental, or lifestyle factors interact to generate observations \citep{Scholkopfetal21, bareinboim2022on, peters2017elements}.
This is particularly important when reasoning about counterfactual scenarios in patients, e.g., \emph{``What would this scan look like in different imaging conditions?''} or \emph{``How would a disease have progressed in this individual under alternative treatments?''}.
Answering such questions requires generative models that can perform isolated edits while preserving patient-specific anatomy.

\citet{pearl2009causality} formalises counterfactual reasoning using structural causal models (SCMs), in which variables are generated by a set of functional assignments, or \emph{mechanisms}, forming a directed acyclic graph.
Counterfactual inference is then performed via a simple three-step procedure: \emph{abduction-action-prediction}.
\citet{pawlowski2020deepscm} implement SCMs with deep generative models to generate counterfactual images, and, in the absence of ground-truth counterfactuals, \citet{monteiro2023axiomatic} evaluate counterfactual soundness using metrics grounded in the axiomatic properties of mechanisms \citep{galles1998axiomatic,halpern2000axiomatizing}.
Building on this, hierarchical VAEs \citep[HVAEs, ][]{vahdat2020nvae, child2020very}, diffusion models \citep{song2021denoising, preechakul2022diffusion, rombach2022high}, and optimal-transport flow matching \citep{lipman2023flow, pmlr-v202-pooladian23a, tong2024improving} have been used to parameterise mechanisms for generating high-fidelity counterfactuals in mammography, 2D and 3D brain MRIs, subcortical meshes and chest X-rays \citep{ribeiro2023high, ribeiro2025counterfactual, rasal2025diffusion, rasal2022deep, xia2025decoupledcfg, peng2025latent}.
However, deploying these models in clinical settings is limited by access to large, annotated datasets \citep{willemink2020preparing,kaissis2020secure}.
As such, bespoke training is impractical, motivating methods that adapt pretrained models for counterfactual inference in low-data regimes.

A natural alternative is image editing with pretrained text-to-image foundation models.
This is analogous to counterfactual inference \citep{zevcevic2022pearl,pan2024counterfactual,pan2025counterfactual} and is increasingly applied to medical image editing \citep{gu2023biomedjourney, perez-garcia_bond-taylor_radedit, kumar2025prism, cooke2025roentmod}.
These methods control images through text prompts, often with spatial annotations such as masks or bounding boxes to localise edits \citep{couairon2023diffedit,lin2024text}.
However, prompt engineering requires significant expertise, and spatial annotations are often unavailable in clinical practice \citep{xing2023less,wang2021annotation,ma2024segment}.
Moreover, masking cannot capture global morphological variations required for edits on attributes such as view, age, or anatomy.
While recent work introduces causal structure into foundation models via adapters \citep{tong2026causaladapter}, the resulting approach ignores distribution shifts and biases that medical images are often subject to \citep{Castro2020-ic,roschewitz2023automatic}.

\begin{figure*}[t]
\centering

\newlength{\teasergap}\setlength{\teasergap}{6pt}%
\newlength{\teaserimg}\setlength{\teaserimg}{\dimexpr(\textwidth-2\teasergap)/8\relax}%
\begin{minipage}[t]{\dimexpr2\teaserimg\relax}\centering

  \teaserDoLabel{\textit{Change sex to female}}\\[2pt]

  \begin{minipage}{0.5\linewidth}\centering\teaserHdr{Observation}\end{minipage}%
  \begin{minipage}{0.5\linewidth}\centering\teaserHdr{\textbf{\textsc{sa-spec}} (Ours)}\end{minipage}\\[2pt]

  \begin{minipage}{0.5\linewidth}{\setlength{\fboxsep}{0pt}\fcolorbox{black}{white}{\teasercell{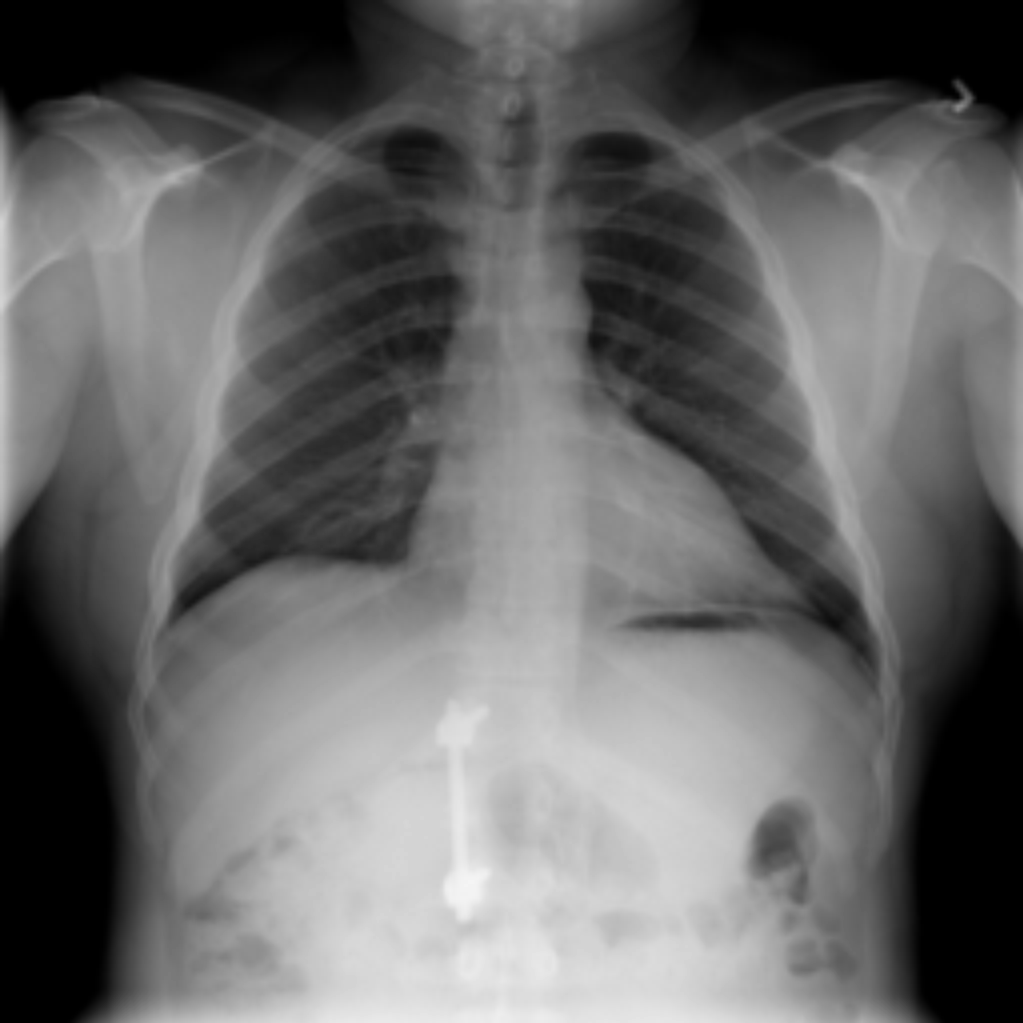}}}\end{minipage}%
  \begin{minipage}{0.5\linewidth}{\setlength{\fboxsep}{0pt}\fcolorbox{black}{white}{\teasercell{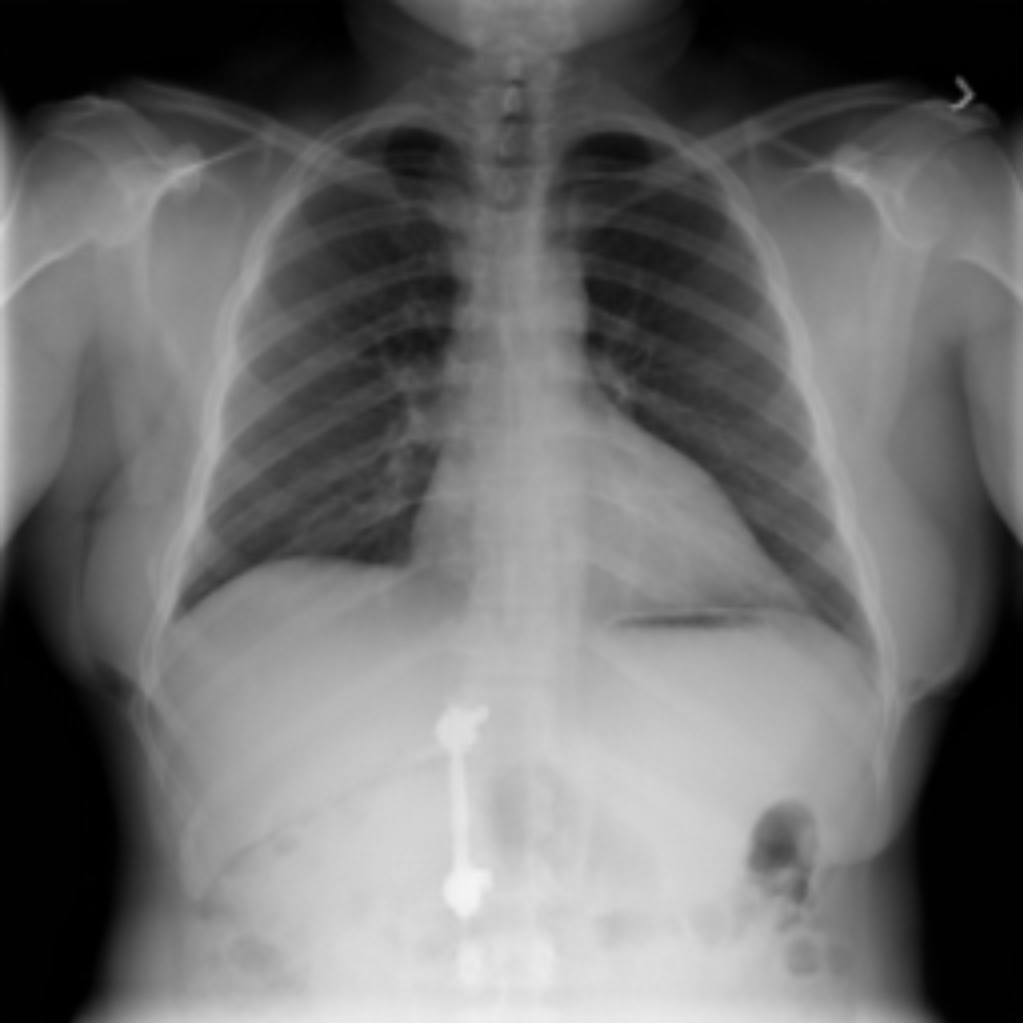}}}\end{minipage}\\[0pt]

  \begin{minipage}[c]{0.5\linewidth}\raggedleft\rotatebox{90}{\scriptsize Direct effect}\hspace*{3pt}\end{minipage}%
  \begin{minipage}[c]{0.5\linewidth}{\setlength{\fboxsep}{0pt}\fcolorbox{gray!20}{white}{\teasercell{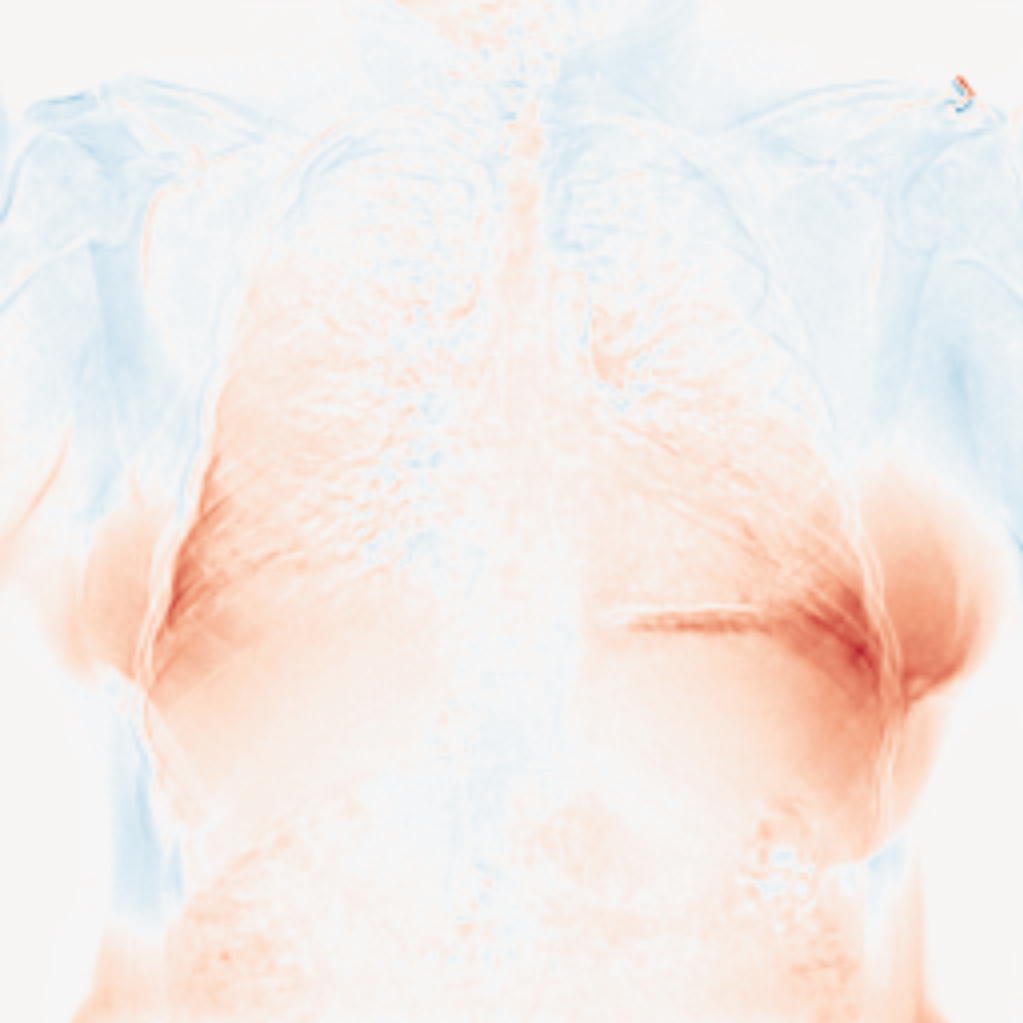}}}\end{minipage}\\[2pt]

  \teaserSubcap{(a) Low-data \spec{sa-spec}\\sex intervention ($\bm{10\%}$).}
\end{minipage}%
\hspace{\teasergap}%
\begin{minipage}[t]{\dimexpr2\teaserimg\relax}\centering

  \teaserDoLabel{\textit{Remove medical device}}\\[2pt]

  \begin{minipage}{0.5\linewidth}\centering\teaserHdr{Observation}\end{minipage}%
  \begin{minipage}{0.5\linewidth}\centering\teaserHdr{\textbf{\textsc{sa-spec}} (Ours)}\end{minipage}\\[2pt]

  \begin{minipage}{0.5\linewidth}{\setlength{\fboxsep}{0pt}\fcolorbox{black}{white}{\teasercell{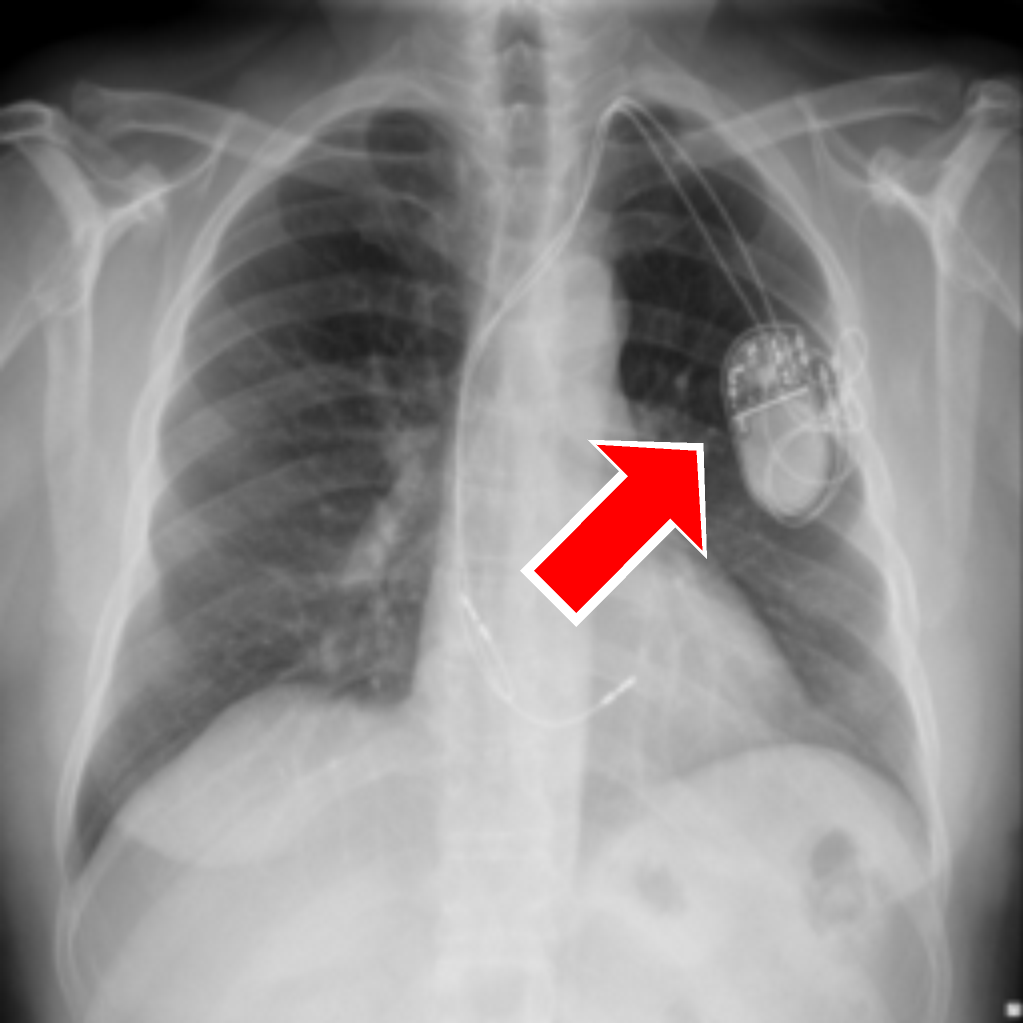}}}\end{minipage}%
  \begin{minipage}{0.5\linewidth}{\setlength{\fboxsep}{0pt}\fcolorbox{black}{white}{\teasercell{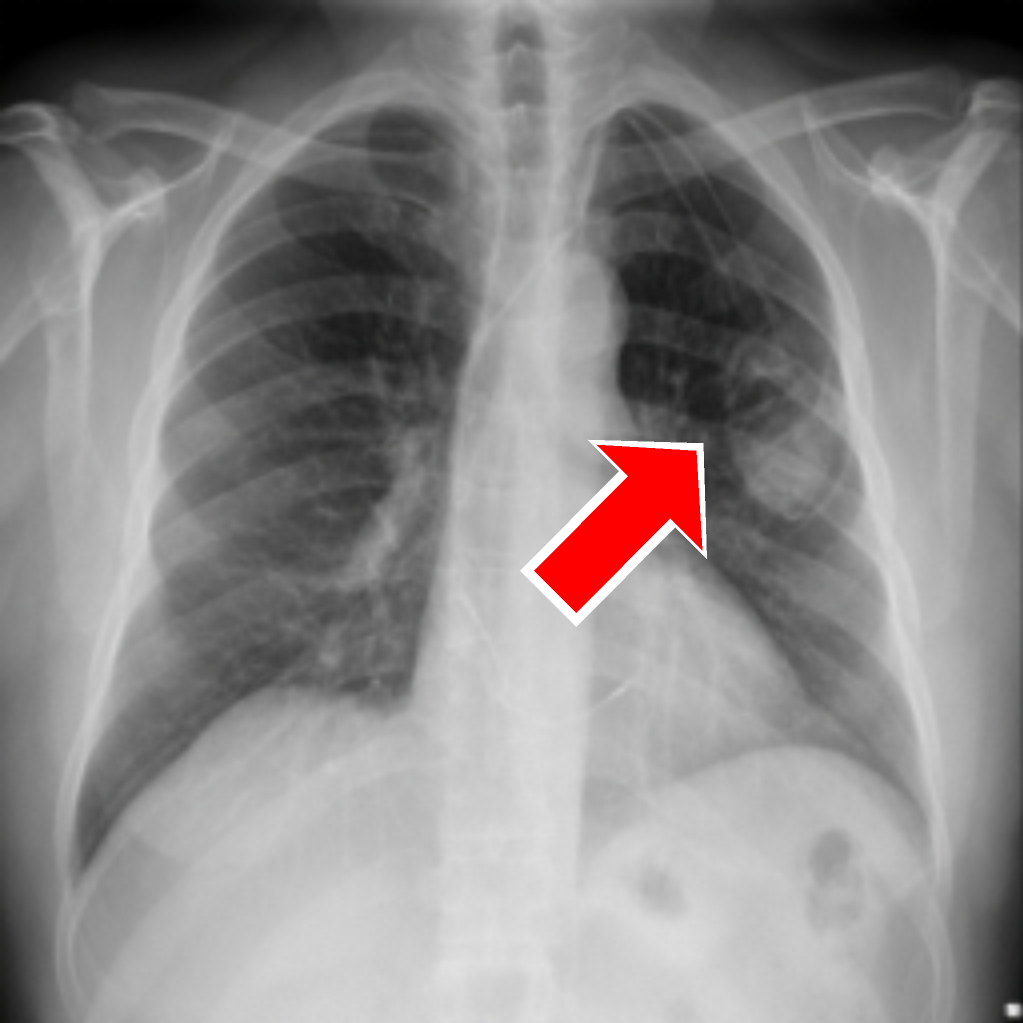}}}\end{minipage}\\[0pt]

  \begin{minipage}[c]{0.5\linewidth}\raggedleft\rotatebox{90}{\scriptsize Direct effect}\hspace*{3pt}\end{minipage}%
  \begin{minipage}[c]{0.5\linewidth}{\setlength{\fboxsep}{0pt}\fcolorbox{gray!20}{white}{\teasercell{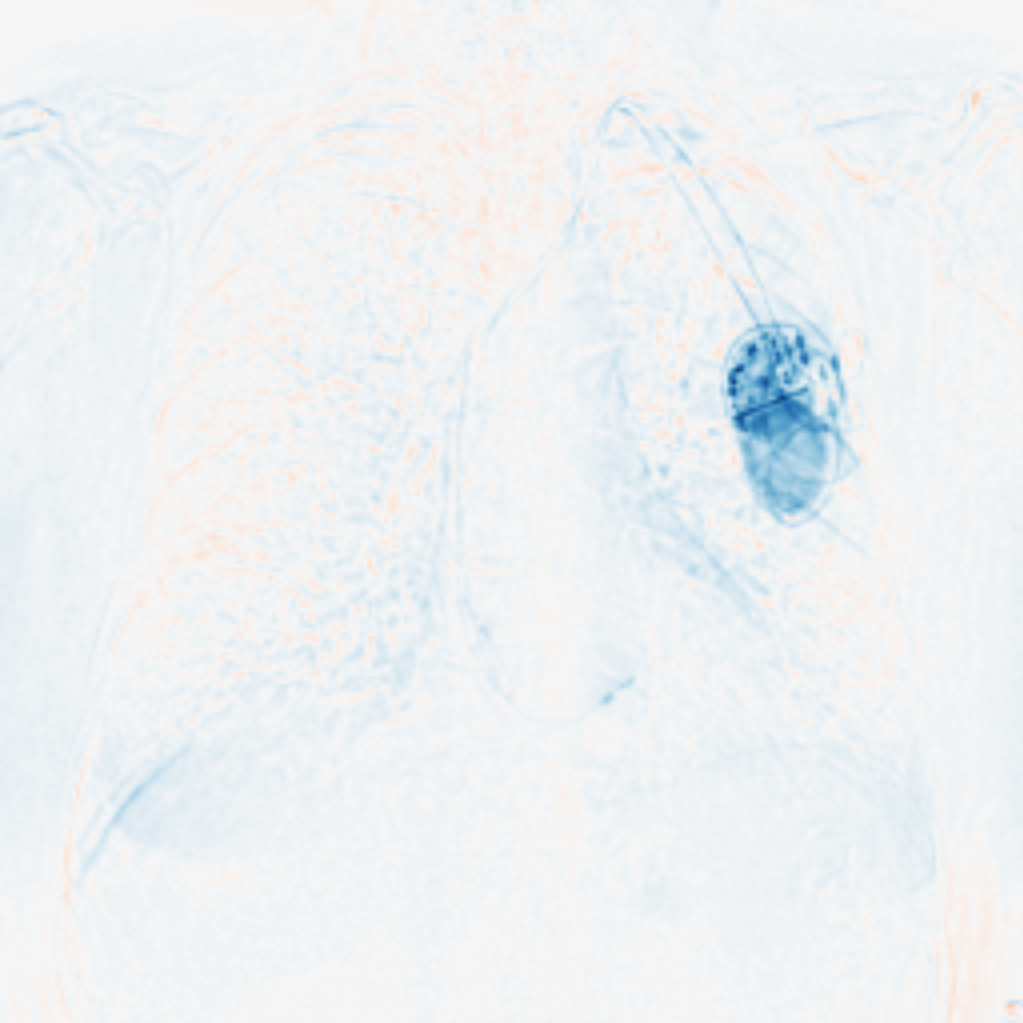}}}\end{minipage}\\[2pt]

  \teaserSubcap{(b) BRAX device removal\\via specialisation.}
\end{minipage}%
\hspace{\teasergap}%
\begin{minipage}[t]{\dimexpr4\teaserimg\relax}\centering

  \teaserDoLabel{\textit{Add pleural effusion disease}}\\[2pt]

  \begin{minipage}{0.25\linewidth}\centering\teaserHdr{Observation}\end{minipage}%
  \begin{minipage}{0.25\linewidth}\centering\teaserHdr{SDXL}\end{minipage}%
  \begin{minipage}{0.25\linewidth}\centering\teaserHdr{RadEdit}\end{minipage}%
  \begin{minipage}{0.25\linewidth}\centering\teaserHdr{\textbf{\textsc{m-spec}} (Ours)}\end{minipage}\\[2pt]

  \begin{minipage}{0.25\linewidth}{\setlength{\fboxsep}{0pt}\fcolorbox{black}{white}{\teasercell{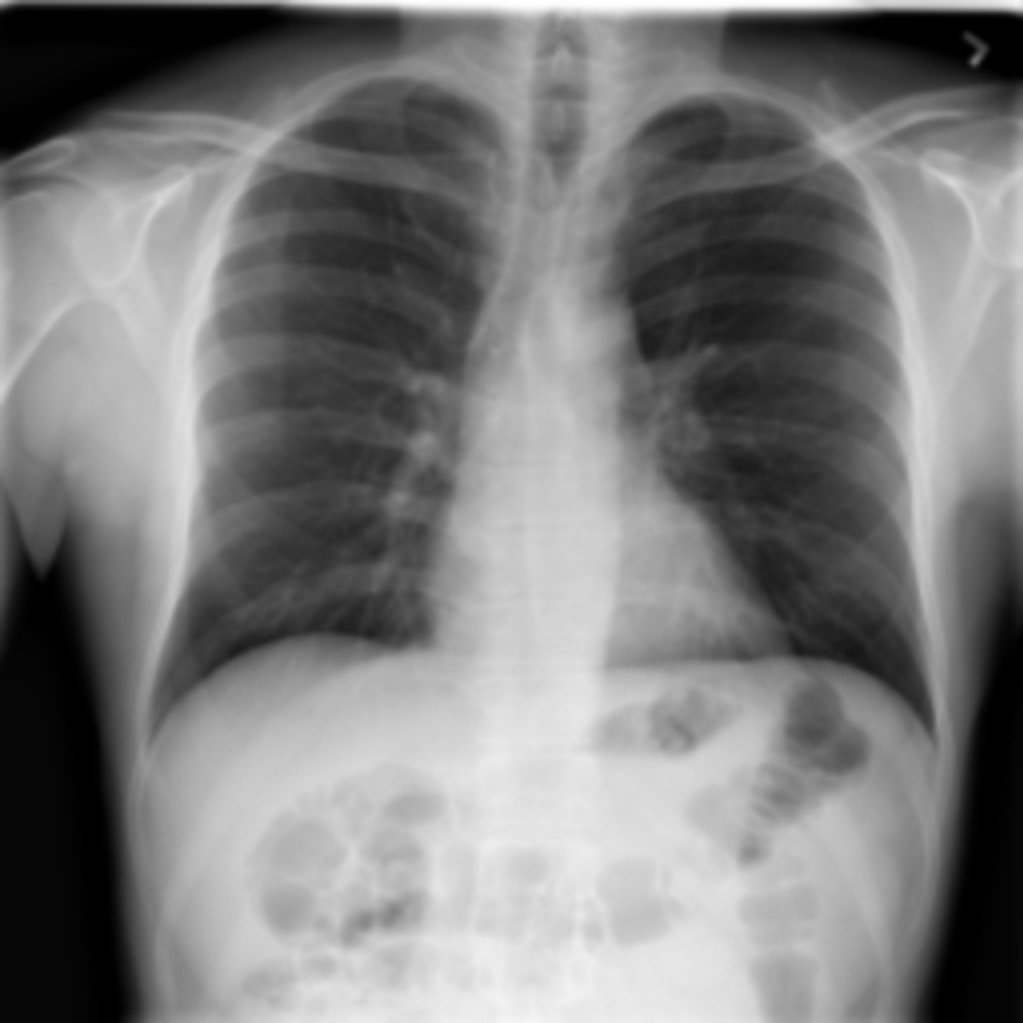}}}\end{minipage}%
  \begin{minipage}{0.25\linewidth}{\setlength{\fboxsep}{0pt}\fcolorbox{black}{white}{\teasercell{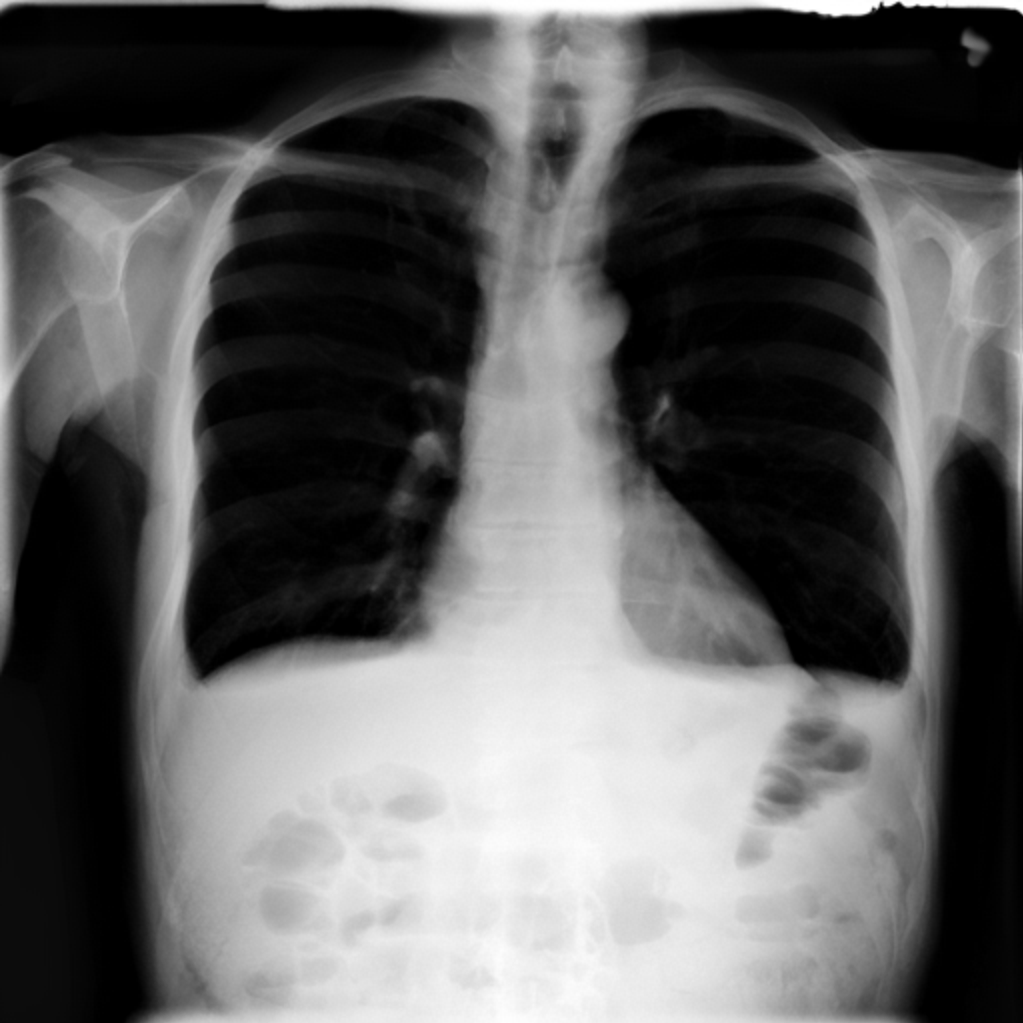}}}\end{minipage}%
  \begin{minipage}{0.25\linewidth}{\setlength{\fboxsep}{0pt}\fcolorbox{black}{white}{\teasercell{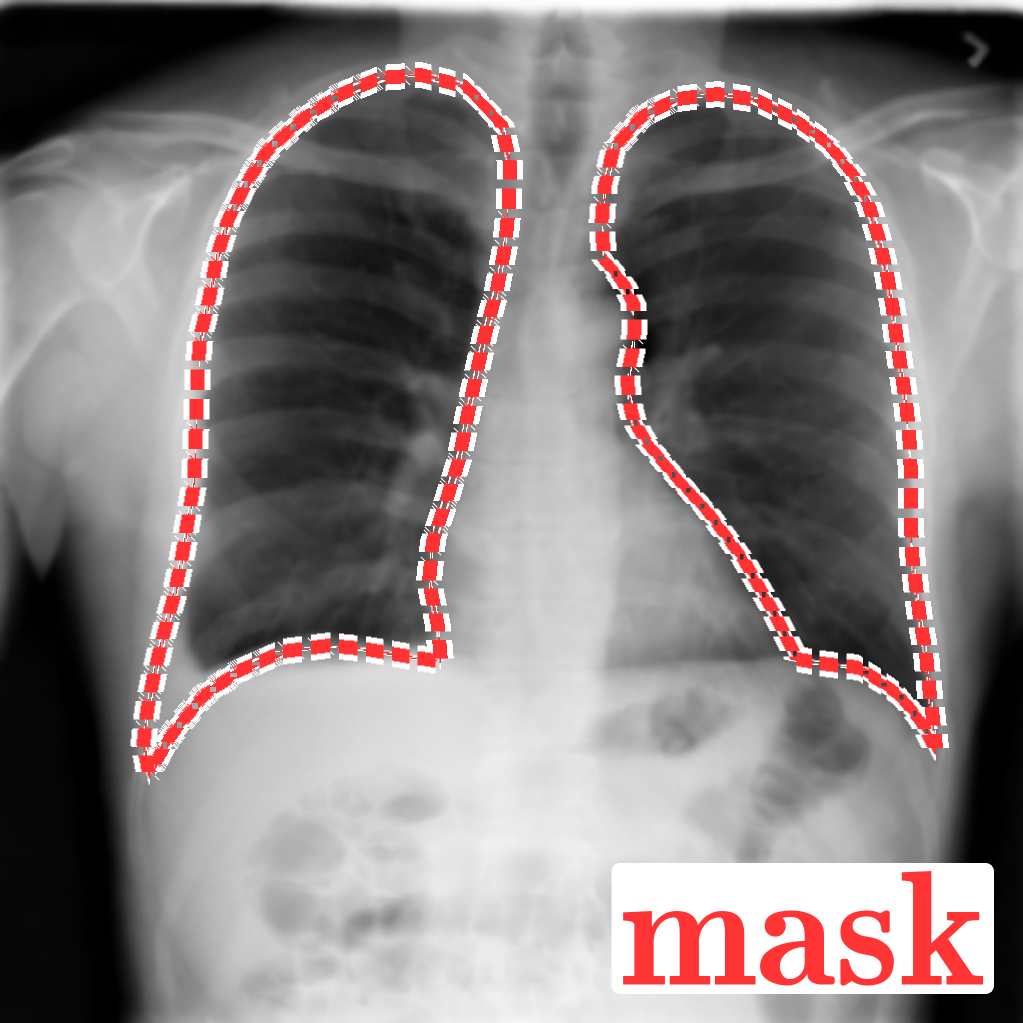}}}\end{minipage}%
  \begin{minipage}{0.25\linewidth}{\setlength{\fboxsep}{0pt}\fcolorbox{black}{white}{\teasercell{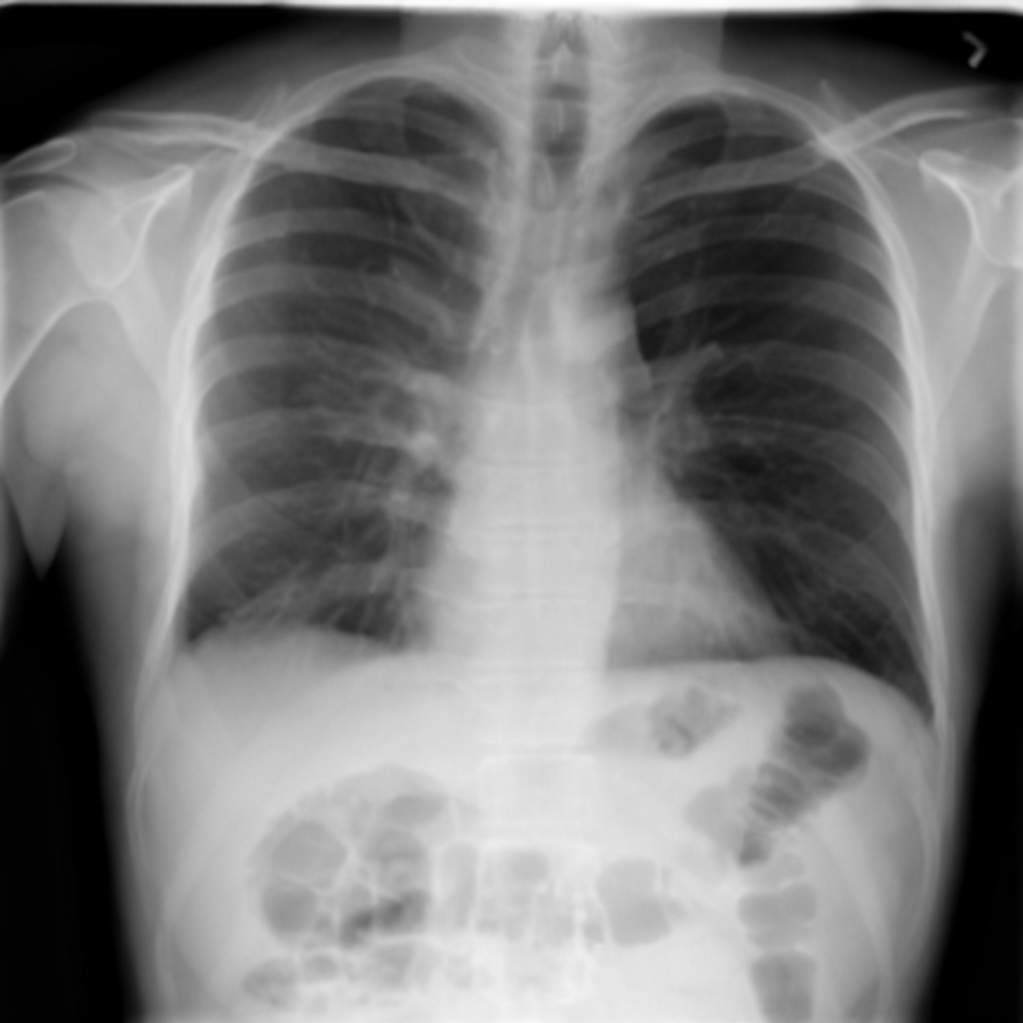}}}\end{minipage}\\[0pt]

  \begin{minipage}[c]{0.25\linewidth}\raggedleft\rotatebox{90}{\scriptsize Direct effect}\hspace*{3pt}\end{minipage}%
  \begin{minipage}[c]{0.25\linewidth}{\setlength{\fboxsep}{0pt}\fcolorbox{gray!20}{white}{\teasercell{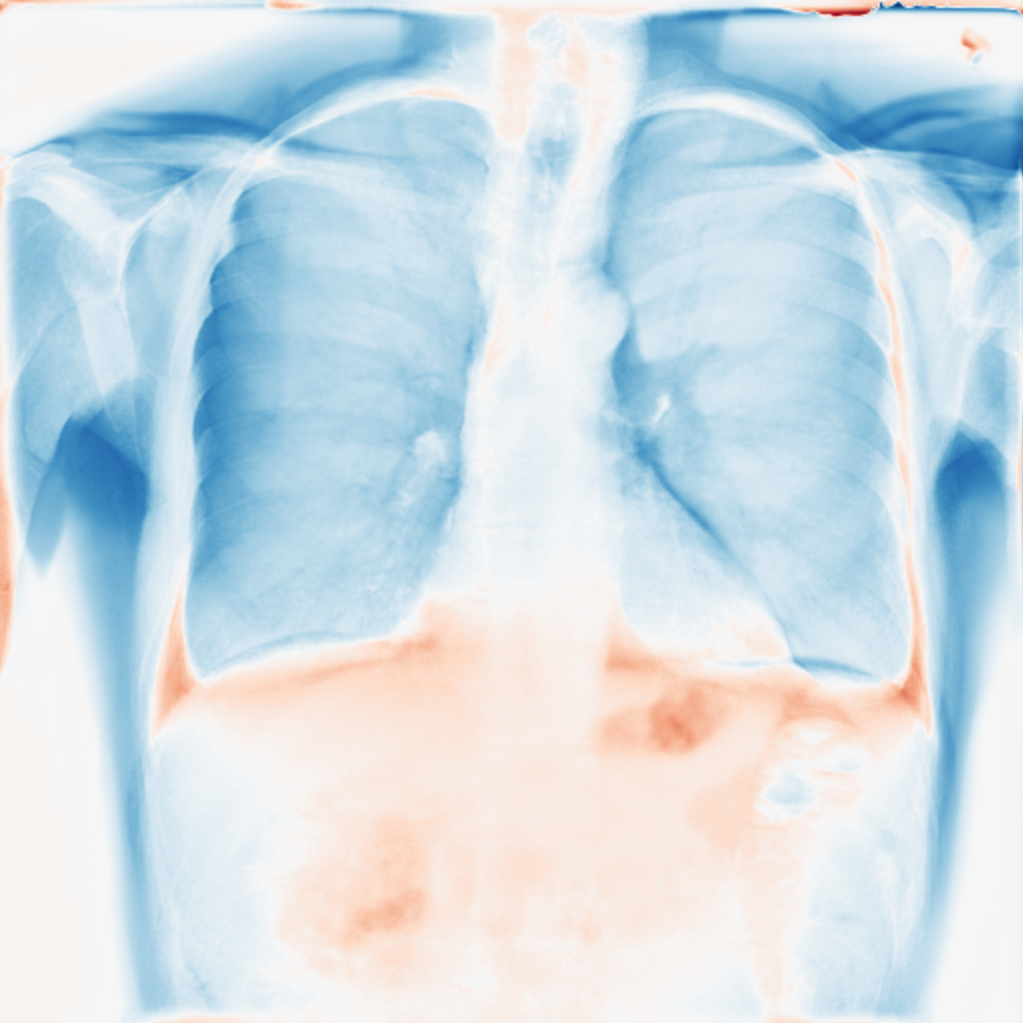}}}\end{minipage}%
  \begin{minipage}{0.25\linewidth}{\setlength{\fboxsep}{0pt}\fcolorbox{gray!20}{white}{\teasercell{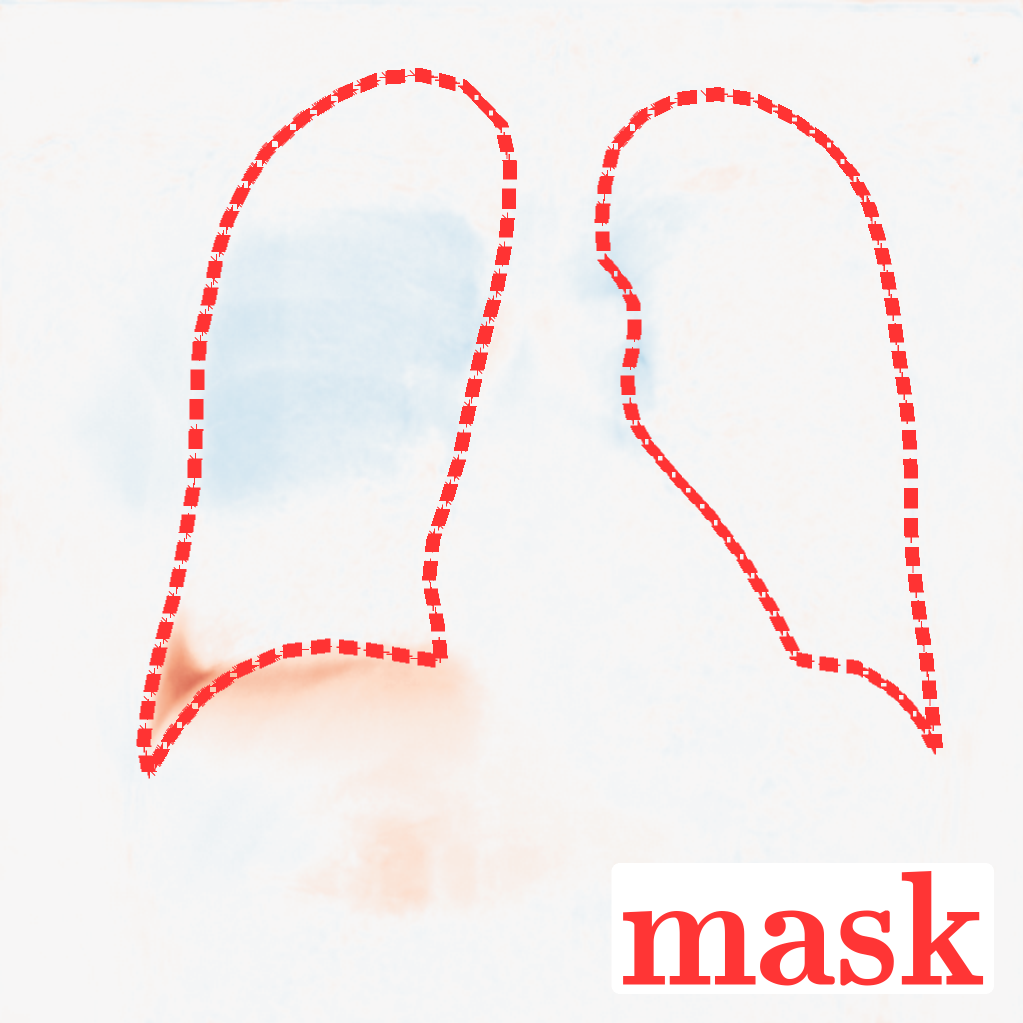}}}\end{minipage}%
  \begin{minipage}{0.25\linewidth}{\setlength{\fboxsep}{0pt}\fcolorbox{gray!20}{white}{\teasercell{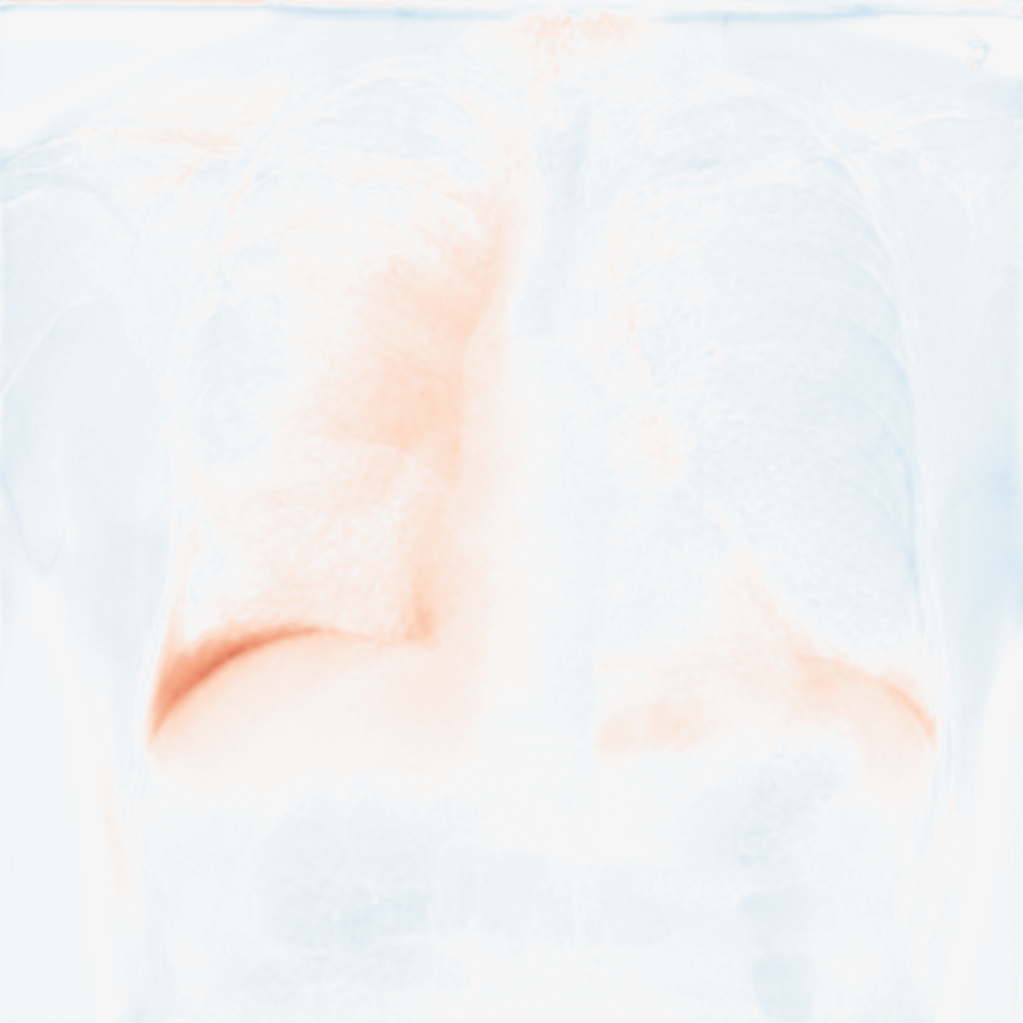}}}\end{minipage}\\[2pt]

  \teaserSubcap{(c) \spec{m-spec} specialises a pretrained text-to-image backbone without requiring masks.}
\end{minipage}

\caption{\textbf{Specialisation enables principled counterfactual generation across three clinical setups, adapting across distribution shift.} \textbf{(a)} \spec{sa-spec} produces controlled sex interventions even with only $\bm{10\%}$ of target-domain training data. \textbf{(b)} \spec{sa-spec} can introduce interventions not modelled during pretraining, like device. \textbf{(c)} \spec{m-spec} adds explicit causal control to a pretrained text-to-image backbone, whereas prior work such as SDXL \citep{sdxl} alters the image globally and RadEdit \citep{perez-garcia_bond-taylor_radedit} requires a spatial mask.}
\label{fig:teaser}
\end{figure*}

We argue that widespread adoption of counterfactual reasoning in the healthcare domain requires adapting foundation models to operate under distribution shift and limited data, as in medical image generation \citep{Bluethgen2024}. In this work, we propose \textbf{\emph{specialisation}}, a framework for adapting pretrained generators to parameterise causal mechanisms under distribution shift in low-data regimes.
We then train RadCF, our method for counterfactual image generation in radiology, and study three specialisation settings (\Cref{fig:teaser}).
Our contributions are as follows:
\begin{itemize}
\itemsep0em
    \item We propose \emph{specialisation}, a data and parameter-efficient framework that adapts pretrained, non-causal generative models into counterfactual image mechanisms (\Cref{subsec:specialisation}). We formulate a mismatch-to-adaptation principle that links structural, acquisition and measurement mismatches to the model components that should be adapted.
    \item We introduce RadCF for counterfactual inference with latent flow matching (\Cref{subsec:flow_mechanism}), which improves axiomatic soundness over prior work on radiology datasets.
    \item We show that \emph{specialisation} extends across dataset shifts (\Cref{sec:result_spec}) and generative model classes (\Cref{sec:result_mspec}), and that the resulting counterfactuals can detect and mitigate shortcut learning in a downstream classifier (\Cref{sec:result_utility}).
\end{itemize}

\noindent\textbf{Relation to prior work (\Cref{app:related_work}).}
We build on three areas:
(1) Deep generative SCMs use VAEs, diffusion models, or flow-based mechanisms for counterfactual image generation\citep{ribeiro2023high, ribeiro2025counterfactual, rasal2025diffusion, rasal2022deep, xia2025decoupledcfg, peng2025latent}, but are typically trained from scratch for each target domain.
This is impractical when large annotated clinical datasets are difficult to obtain \citep{willemink2020preparing, kaissis2020secure}.
Unlike image-space flows \citep{ribeiro2025counterfactual}, we operate in latent space for high-resolution generation.
(2) Generative foundation models provide high-quality edits through prompts, masks, or bounding boxes \citep{gu2023biomedjourney,perez-garcia_bond-taylor_radedit, kumar2025prism, cooke2025roentmod}, but offer associative rather than interventional control over structured parent variables.
(3) Methods for adding conditioning and efficiently adapting pretrained generators, including ControlNet and LoRA \citep{zhang2023adding,hu2022lora}, provide the technical tools for specialisation.
Causal-Adapter \citep{tong2026causaladapter} adapts a frozen text-to-image backbone for causal inference, but does not study distribution shift.

\section{Background}
\label{sec:background}
Our work generates counterfactuals following the methodology in \citep{pearl2009causality} and is implemented using flow matching models \citep{lipman2023flow}. We briefly review these two topics below.

\subsection{Counterfactual inference}
\label{subsec:scms}

\citet{pearl2009causality} refers to structural causal models (SCMs) to describe causal relationships and compute causal quantities within a system.
We consider acyclic Markovian SCMs $\mathfrak{S}:=(\mathbf{U},\mathbf{X},\mathbf{F})$, with a set of exogenous variables $\mathbf{U} = \{\uu_k\}^{K}_{k = 1}$ which are mutually independent, a set of observed variables $\mathbf{X} = \{\x_k\}^{K}_{k=1}$, and a set of deterministic functions $\mathbf{F}=\{f_k\}_{k=1}^{K}$ called \emph{mechanisms} defining causal relationships via acyclic assignments as
\begin{equation}
\label{eq:scm_mechanism}
    \x_k=f_k(\pa_k,\uu_k),
    \qquad
    \pa_k\subseteq\mathbf{X}\setminus\{\x_k\},
    \qquad
    k\in\{1,\ldots,K\},
\end{equation}
where $\pa_k$ are the direct causes, or \emph{parents}, of $\x_k$.

Counterfactual inference follows a three-step procedure \citep{pearl2009causality}: \emph{abduction} infers the exogenous posterior given the factual observations; \emph{action} replaces the structural assignment of the intervened variable, denoted $\doo(\cdot)$; and \emph{prediction} evaluates the modified SCM at the counterfactual parents $\pacf_k$, retaining the same abducted exogenous states.
For mechanisms that are bijective in $\uu_k$ at fixed $\pa_k$, abduction reduces to inversion, $\uu_k=f_k^{-1}(\x_k,\pa_k)$.
For non-invertible mechanisms, it instead requires the posterior over exogenous states compatible with the factual observation.

Counterfactuals from correct SCMs satisfy axiomatic properties \citep{galles1998axiomatic,halpern2000axiomatizing}.
\citet{monteiro2023axiomatic} operationalise these as metrics for \emph{composition}, \emph{reversibility}, and \emph{effectiveness}.
\emph{Composition} measures consistency under null interventions ($\doo(\pa)$) and \emph{reversibility} evaluates cycle-consistency ($\doo(\pacf) \rightarrow \doo(\pa)$), scored by LPIPS \citep{zhang2018lpips}.
Together, they indicate a mechanism's ability to preserve identity in counterfactuals.
Finally, \emph{effectiveness} measures faithfulness to interventions via pseudo-oracle classifiers $p(\pa | \x)$ to verify that target attributes change and non-target attributes are preserved (e.g., sex, age), scored by accuracy and MAE.
This evaluation strategy is standard for counterfactuals \citep{rasal2025diffusion,ribeiro2025counterfactual,roschewitz2023automatic}; all metrics, pseudo-oracle training, confounder mitigation and sensitivity analysis are detailed in \Cref{appendix:metricImplementations}.
\citet{melistas2025benchmarking} further propose \emph{realism} for counterfactual image fidelity, which we score by KID \citep{binkowski2018demystifying}.


\subsection{Flow matching}
\label{subsec:flows}
Flow matching models \citep{lipman2023flow} learn transport dynamics from a base to a data distribution to generate samples.
\citet{ma2024sit} define a linear probability path $\z_t = (1-t)\z_0 + t\z_1$, $t\in[0,1]$, in the latent space of a VAE \citep{rombach2022high,esser2020taming}, interpolating between the data encoding $\z_0 = E_{\phi_1}(\x_0)$ at $t=0$ and noise $\z_1\sim\mathcal{N}(0,\identity)$ at $t=1$, with decoder $\hat{\x}_0 = D_{\phi_2}(\z_0)$.
Its time derivative is $\z_1 - \z_0$, which the conditional velocity field $\mathbf{v}_{\theta}(\z_t, t, \cc)$ is trained to approximate,
\begin{equation}
    \mathcal{L}_{\text{FM}}(\theta)
    = \mathbb{E}_{t \sim \mathcal{U}[0, 1], \z_1 \sim \mathcal{N}(0, \identity), \z_0, \cc}
    \left \lVert
    \mathbf{v}_\theta(\z_t, t, \cc) - (\z_1 - \z_0)
    \right \rVert_2^2.
    \label{eq:fm_loss}
\end{equation}
When the encoder and decoder are trained independently of the velocity field $\mathbf{v}_\theta(\cdot)$ \citep{rombach2022high,Peebles2022DiT}, mismatches between the latent representation and the learned transport can degrade generative quality.
The REPA-E framework \citep{leng2025repae} jointly optimises the velocity field $\theta$ and the VAE parameters $\phi = \{ \phi_1, \phi_2 \}$, aligning them to DINOv2 representations \citep{oquab2024dinov2learningrobustvisual} for efficient training \citep{yu2025repa}.
The full objective is
\begin{equation}
\mathcal{L}_{\text{REPA-E}}(\theta,\phi,\xi)
=
\mathcal{L}_{\text{FM}}(\theta)
+
\lambda \mathcal{L}_{\text{align}}(\theta,\phi,\xi)
+
\eta \mathcal{L}_{\text{reg}}(\phi),
\label{eq:repae}
\end{equation}
where $\mathcal{L}_{\text{align}}(\cdot)$ is the alignment term, $\mathcal{L}_{\text{reg}}(\cdot)$ learns the VAE with perceptual losses (e.g.\ MSE, LPIPS) and a KL-divergence over the prior.
The coefficients $\lambda$ and $\eta$ are hyperparameters.
Alignment matches intermediate transport features to frozen DINOv2 features \citep{oquab2024dinov2learningrobustvisual} through a projection head $g_\xi$.

We solve the probability-flow ODE $\frac{d\z_t}{dt} = \mathbf{v}_\theta(\z_t, t, \cc)$ backward in time for generation ($\z_0$) and forward in time for inversion ($\z_1$),
\begin{equation}
\label{eq:gen_inv_flow}
\begin{alignedat}{2}
\z_0 &:= T_\theta(\z_1, \cc)
= \z_1 - \int_0^1 \mathbf{v}_\theta(\z_t, t, \cc)\, dt,
\qquad&
\z_1 &:= T_\theta^{-1}(\z_0, \cc)
= \z_0 + \int_0^1 \mathbf{v}_\theta(\z_t, t, \cc)\, dt .
\end{alignedat}
\end{equation}
where $T_\theta(\cdot)$ defines a bijective flow under regularity assumptions on $\mathbf{v}_\theta(\cdot)$ \citep{chen2018neuralode,albergo2023stochastic,ribeiro2025counterfactual}, commonly implemented via the Euler sampler \citep{chen2018neuralode,song2021scorebased,lipman2023flow,ma2024sit}.

\section{Method: specialisation for counterfactual image mechanisms}
\label{sec:method}

In medical settings, imaging datasets are typically small and cannot easily be pooled due to privacy constraints \citep{willemink2020preparing,kaissis2020secure}.
Since existing counterfactual generative methods are trained from scratch, these data challenges limit their use in the healthcare domain.
Additionally, observed distributions vary due to acquisition shifts in $\x$ arising from differences in scanners and imaging protocols, and measurement bias in $\pa$ where observed labels act as imperfect proxies for underlying clinical factors \citep{Castro2020-ic,roschewitz2023automatic}.
This limits cross-domain generalisability of pretrained models.

\subsection{Specialisation}
\label{subsec:specialisation}

To address these gaps, we introduce \emph{specialisation}, a parameter-efficient procedure that adapts a pretrained generative model into a counterfactual image mechanism.
We consider a generator pretrained on $p_s(\x)$ or $p_s(\x\mid\cc_s)$, where $\cc_s$ denotes the source conditioning variables.
The target data follow $p_t(\x,\pa)$, where $\pa$ denotes the image parents specified by the target SCM.
Specialisation reuses the source-pretrained image prior while adapting the generator to the target image domain and parent structure.
Here, we summarise three possible mismatch scenarios following \citet{Castro2020-ic}, and the corresponding specialisation strategy.


\noindent\textbf{Structural mismatch and structural specialisation (\spec{s-spec}).}
Structural mismatch occurs when the source conditioning structure does not expose the target parents $\pa$.
It includes unconditional generators, models conditioned on source variables $\cc_s$ that omit $\pa$ or cannot be mapped to it, and target tasks that introduce parents absent at pretraining.
Such models may encode image variation associated with $\pa$, but they cannot evaluate the same observation-specific mechanism at factual and counterfactual parent values while holding $\uu$ fixed.
We therefore propose \spec{s-spec}, which introduces the target parent inputs and adapts the generator to evaluate the image mechanism at factual and counterfactual parent assignments.

\noindent\textbf{Acquisition mismatch and structural-acquisition specialisation (\spec{sa-spec}).}
Acquisition mismatch occurs when source and target datasets differ in scanners, imaging protocols, resolution, or image processing, so that the same underlying anatomy is represented differently in the observed images \citep{Castro2020-ic}.
These differences can impair reconstruction of target observations even after the generator has learned to respond to the target parents.
\spec{sa-spec} therefore extends \spec{s-spec} by adapting the image representation together with the parent-dependent generation.
This allows the specialised mechanism to accommodate target-domain appearance while retaining the parent-dependent image structure.

\noindent\textbf{Measurement mismatch and measurement specialisation (\spec{m-spec}).}
Measurement mismatch occurs when the source model conditions on a proxy for $\pa$ rather than on the target parents themselves.
In text-conditioned medical generators, free text may omit or conflate target variables or encode information outside the target SCM \citep{kumar2025prism,cooke2025roentmod,Castro2020-ic,Hernan2009,Shahar2009CausalDF}.
Prompt-driven editors also express interventions through selected wording and often use spatial annotations to localise changes \citep{perez-garcia_bond-taylor_radedit}.
\spec{m-spec} bypasses the text encoder and maps structured parents to the backbone's native conditioning interface.
This preserves the source model's text-conditioned image knowledge while replacing an indirect proxy with an explicit intervention interface.
Causal-Adapter \citep{tong2026causaladapter} addresses this same measurement mismatch with a different adapter design; \Cref{sec:result_mspec} compares the two on a shared pretrained text-to-image backbone.

\subsection{RadCF: a latent flow mechanism for specialisation}
\label{subsec:flow_mechanism}
Our image mechanism uses the reverse latent flow in \Cref{eq:gen_inv_flow} for generation and its forward direction for abduction,
\begin{equation}
    \x := f_{\psi}(\uu, \pa) \approx D_{\phi_2}\!\left(T_\theta(\uu, \pa)\right),
    \qquad
    \uu := f^{-1}_{\psi}(\x, \pa) \approx T_\theta^{-1}\!\left(E_{\phi_1}(\x), \pa\right),
    \label{eq:flow_mechanism}
\end{equation}
with parameters $\psi = \{\phi_1, \phi_2, \theta\}$, and $p(\uu) = p(\z_1) = \mathcal{N}(0, \identity)$ defining the exogenous prior as the marginal distribution at the endpoint of the flow.
The parents $\pa$ are passed into $T_\theta(\cdot)$ via a learnable embedding $g_{\omega}(\pa) = \sum_{\pa_k \in \pa} g_{\omega_k}(\pa_k)$ added to the timestep embedding, akin to the metadata conditioning pattern in \citep{khanna2024diffusionsat} and detailed in \Cref{appendix:model_architecture}, so $T_\theta(\cdot)$ can be learned on observational data using the REPA-E objective in \Cref{eq:repae} with $\cc = \pa$.
Operating in a semantically compressed latent space, rather than the pixel space of the \emph{spatial mechanisms} of \citet{rasal2025diffusion}, additionally yields higher-resolution, higher-fidelity counterfactuals.

\noindent\textbf{Identity preservation for latent flow.}
$T_{\theta}(\cdot)$ is a continuous-time PF-ODE, which is bijective under standard assumptions \citep{chen2018neuralode,albergo2023stochastic}, and reaches its Gaussian endpoint at finite $t=1$.
Diffusion mechanisms instead learn a discrete-time score function and approach the Gaussian only as $t\rightarrow\infty$ \citep{rasal2025diffusion,komanduri2024causaldiffae,peng2025latent,kingma2021on}, compounding approximation error across abduction and generation and leaving residual information about $\x$ in $\uu$; our abducted $\uu$ is correspondingly less entangled with $\pa$ (\Cref{appendix:probe}).
Unlike the identifiable pixel-space flow of \citet{ribeiro2025counterfactual}, our bijectivity holds only in $\z_0$, so detail discarded by $E_{\phi_1}(\cdot)$ is unrecoverable; REPA-E limits this by learning $\phi$ and $\theta$ jointly rather than fixing an autoencoder the flow must accommodate \citep{tong2026causaladapter,peng2025latent}.
We therefore trade image-space identifiability for a transport problem posed over semantics rather than pixels, at a cost in composition and reversibility that \Cref{tab:main_results} quantifies.

\subsection{Specialisation objectives}
\label{subsec:specialisation_objectives}

The diagnosed mismatch determines which parameters are updated during specialisation.

\spec{s-spec} adapts the pretrained velocity field to use $\pa$ via Low-Rank Adaptation \citep[LoRA, ][]{hu2022lora}:
\begin{equation}
\begin{aligned}
\mathcal{L}_{\spec{s-spec}}
    (\Delta\theta', \omega, \xi; \phi)
&=
\mathbb{E}_{t,\epsilon,\mathbf{z}_0,\mathbf{pa}}
\left\|
v_{\theta'+\Delta\theta'}
    \bigl(\mathbf{z}_t,t,g_\omega(\mathbf{pa})\bigr)
-
(\epsilon-\mathbf{z}_0)
\right\|_2^2
\\
&\quad+
\lambda\mathcal{L}_{\mathrm{align}}
    (\theta'+\Delta\theta',\phi,\xi,\omega).
\end{aligned}
\label{eq:s_spec_equation}
\end{equation}
Here, $\z_0=E_{\phi_1}(\x_0)$, $\epsilon=\z_1\sim\mathcal{N}(0,\identity)$, and $\z_t=(1-t)\z_0+t\epsilon$.
The VAE parameters $\phi$ are held fixed at their source-pretrained values.
This yields the conditional model $T_{\theta}(\uu,\pa)$ with $\theta = \theta' + \Delta\theta'$, where the low-rank update applies to every linear layer of $\mathbf{v}_{\theta'}(\cdot)$.
We provide the ablation results with rank and adapter ablations in \Cref{appendix:ablations}

Our mechanism decouples components responsible for composition (reconstruction) and effectiveness (intervention faithfulness); domain-specific pixel-level details are captured by a VAE and semantic control is governed by the latent flow.
\spec{sa-spec} jointly adapts the VAE parameters $\phi=\{\phi_1,\phi_2\}$ and the low-rank flow update $\Delta\theta'$.
Following \Cref{eq:repae}, we extend the \spec{s-spec} as below:
\begin{equation}
\mathcal{L}_{\spec{sa-spec}}(\Delta\theta',\phi,\omega,\xi)
= \mathcal{L}_{\spec{s-spec}}(\Delta\theta',\omega,\xi;\phi) + \eta\mathcal{L}_{\text{reg}}(\phi).
\label{eq:sa_spec_equation}
\end{equation}

\spec{m-spec} replaces text conditioning with $g_\omega(\pa)$ using \Cref{eq:s_spec_equation}, thereby obtaining explicit control over $\pa$ for text-conditioned models.
Since the velocity field $\mathbf{v}_{\theta'}(\cdot)$ can be reparameterised as $\bm{\epsilon}_{\theta'}(\z_t, t, \pa) = \sigma_t \z_t + \alpha_t \mathbf{v}_{\theta'}(\z_t, t, \pa)$, \spec{m-spec} generalises to latent diffusion models,
\begin{equation}
    \mathcal{L}_{\spec{m-spec}}(\Delta \theta', \omega)
        = \mathbb{E}_{t, \eps, \z_0, \pa}
        \lVert \boldsymbol{\epsilon}_{\theta' + \Delta \theta'}(\z_t, t, g_\omega(\pa)) - \eps \rVert_2^2.
    \label{eq:m_spec_diffusion}
\end{equation}

\section{Results}
\label{sec:results}
We evaluate specialisation as a data and parameter-efficient adaptation strategy (\Cref{sec:result_spec}), \spec{m-spec} as an extension to text-conditioned editors (\Cref{sec:result_mspec})
and demonstrate the utility of our counterfactuals on a downstream use case to complement the axiomatic evaluations (\Cref{sec:result_utility}).

Unless otherwise stated, experiments specialise a CheXpert-pretrained model~\citep{irvin2019chexpert} to NIH-14~\citep{wang2017chestx} using \spec{sa-spec}.
Across all datasets we extract four parent variables from structured metadata: view position $v \in \{\text{AP}, \text{PA}\}$, sex $s \in \{\text{M}, \text{F}\}$, age $a \in [0,1]$ and effusion status $e \in \{0,1\}$.
BRAX~\citep{brax} additionally provides a device label $d \in \{0,1\}$, and presents distinct acquisition and population shifts.
See \Cref{appendix:experiment_details} for experiment details.

For demographic variables such as age and sex, $\doo(\cdot)$ denotes a counterfactual appearance query: ``What would this radiograph look like if the patient were older / of a different sex?'' rather than a physical clinical action.
We acknowledge the ethical limitations of such interventions, although they are valuable for estimating counterfactual fairness \citep{kusner2017}.

\subsection{Specialisation under target-domain shift and data scarcity}
\label{sec:result_spec}


\noindent\textbf{RadCF as a sound counterfactual mechanism.}
We first establish the soundness of latent flow matching against three mechanisms: an HVAE \citep{ribeiro2023high}, a latent-diffusion implementation of the spatial mechanism of \citet{rasal2025diffusion}, and the image-space flow of \citet{ribeiro2025counterfactual}.
Effectiveness and identity preservation (the latter measured by composition and reversibility) often trade off in counterfactual image generation~\citep{rasal2025diffusion,monteiro2023axiomatic}.
\Cref{tab:main_results} shows that RadCF provides the strongest overall balance between them, supporting latent flow matching as a strong foundation for axiomatically sound, high-fidelity counterfactual generation.


\begin{table*}[t]
\centering
\caption{\textbf{RadCF produces more axiomatically sound counterfactuals than baselines.}
Each row block reports one intervention.
Effectiveness is measured by pseudo-oracles.
For the intervened attribute, the corresponding pseudo-oracle metric measures target success. For non-intervened attributes, the metric measures preservation of the original.
KID is scaled by $\times 10^{2}$.
Best per column metrics are in bold.}
\label{tab:main_results}
\small
\setlength{\tabcolsep}{4pt}
\resizebox{\textwidth}{!}{%
\begin{tabular}{ll cccc c c c}
\toprule
 & & \multicolumn{4}{c}{Effectiveness} & Composition & Reversibility & Realism \\
\cmidrule(lr){3-6}
\cmidrule(lr){7-7}
\cmidrule(lr){8-8}
\cmidrule(lr){9-9}
Interv. & Method & $\text{Acc}(v)\uparrow$ & $\text{Acc}(e)\uparrow$ & $\text{Acc}(s)\uparrow$ & $\text{MAE}(a)\downarrow$ & LPIPS\,$\downarrow$ & LPIPS\,$\downarrow$ & KID\,$\downarrow$ \\
\midrule
 \multirow{4}{*}{$\doo(v)$}
 & HVAE & 0.976$_{\,(0.003)}$ & 0.706$_{\,(0.008)}$ & 0.734$_{\,(0.008)}$ & 0.085$_{\,(0.067)}$ & 0.045$_{\,(0.011)}$ & 0.091$_{\,(0.050)}$ & 7.01$_{\,(0.62)}$ \\

 & Diffusion & 0.957$_{\,(0.004)}$ & 0.759$_{\,(0.008)}$ & \textbf{0.953}$_{\,(0.004)}$ & 0.073$_{\,(0.060)}$ & 0.032$_{\,(0.032)}$ & 0.067$_{\,(0.054)}$ & 3.02$_{\,(0.55)}$ \\

 & Image-space flow & 0.984$_{\,(0.002)}$ & \textbf{0.762}$_{\,(0.008)}$ & 0.931$_{\,(0.005)}$ & 0.076$_{\,(0.061)}$ & \textbf{0.011}$_{\,(0.016)}$ & \textbf{0.048}$_{\,(0.039)}$ & 1.98$_{\,(0.58)}$ \\

 & RadCF & \textbf{0.996}$_{\,(0.001)}$ & 0.726$_{\,(0.008)}$ & 0.948$_{\,(0.004)}$ & \textbf{0.060}$_{\,(0.049)}$ & 0.017$_{\,(0.029)}$ & 0.116$_{\,(0.091)}$ & \textbf{0.53}$_{\,(0.26)}$ \\
\midrule
 \multirow{4}{*}{$\doo(e)$}
 & HVAE & 0.971$_{\,(0.003)}$ & \textbf{0.991}$_{\,(0.002)}$ & 0.771$_{\,(0.007)}$ & 0.147$_{\,(0.104)}$ & 0.045$_{\,(0.011)}$ & 0.246$_{\,(0.029)}$ & 12.82$_{\,(1.28)}$ \\

 & Diffusion & \textbf{0.998}$_{\,(0.001)}$ & 0.338$_{\,(0.008)}$ & 0.977$_{\,(0.003)}$ & \textbf{0.056}$_{\,(0.048)}$ & 0.032$_{\,(0.032)}$ & 0.058$_{\,(0.048)}$ & 1.61$_{\,(0.34)}$ \\

 & Image-space flow & 0.996$_{\,(0.001)}$ & 0.497$_{\,(0.009)}$ & \textbf{0.977}$_{\,(0.003)}$ & 0.058$_{\,(0.050)}$ & \textbf{0.011}$_{\,(0.016)}$ & \textbf{0.034}$_{\,(0.029)}$ & 1.24$_{\,(0.41)}$ \\

 & RadCF & 0.997$_{\,(0.001)}$ & 0.659$_{\,(0.009)}$ & 0.968$_{\,(0.003)}$ & 0.058$_{\,(0.049)}$ & 0.017$_{\,(0.029)}$ & 0.039$_{\,(0.067)}$ & \textbf{0.83}$_{\,(0.28)}$ \\
\midrule
 \multirow{4}{*}{$\doo(s)$}
 & HVAE & 0.895$_{\,(0.005)}$ & 0.665$_{\,(0.008)}$ & \textbf{0.981}$_{\,(0.002)}$ & 0.094$_{\,(0.074)}$ & 0.045$_{\,(0.011)}$ & 0.135$_{\,(0.037)}$ & 5.44$_{\,(0.70)}$ \\

 & Diffusion & 0.986$_{\,(0.002)}$ & 0.854$_{\,(0.006)}$ & 0.743$_{\,(0.008)}$ & 0.062$_{\,(0.051)}$ & 0.032$_{\,(0.032)}$ & 0.072$_{\,(0.056)}$ & 1.71$_{\,(0.44)}$ \\

 & Image-space flow & \textbf{0.998}$_{\,(0.001)}$ & \textbf{0.887}$_{\,(0.006)}$ & 0.818$_{\,(0.007)}$ & 0.060$_{\,(0.050)}$ & \textbf{0.011}$_{\,(0.016)}$ & \textbf{0.037}$_{\,(0.034)}$ & 0.92$_{\,(0.34)}$ \\

 & RadCF & 0.997$_{\,(0.001)}$ & 0.807$_{\,(0.007)}$ & 0.964$_{\,(0.003)}$ & \textbf{0.057}$_{\,(0.047)}$ & 0.017$_{\,(0.029)}$ & 0.073$_{\,(0.081)}$ & \textbf{0.48}$_{\,(0.21)}$ \\
\midrule
 \multirow{4}{*}{$\doo(a)$}
 & HVAE & \textbf{1.000}$_{\,(0.000)}$ & \textbf{0.958}$_{\,(0.004)}$ & \textbf{0.989}$_{\,(0.002)}$ & 0.225$_{\,(0.132)}$ & 0.045$_{\,(0.011)}$ & \textbf{0.000}$_{\,(0.003)}$ & 0.74$_{\,(0.27)}$ \\

 & Diffusion & 0.994$_{\,(0.001)}$ & 0.875$_{\,(0.006)}$ & 0.932$_{\,(0.004)}$ & 0.155$_{\,(0.095)}$ & 0.032$_{\,(0.030)}$ & 0.059$_{\,(0.049)}$ & 1.73$_{\,(0.34)}$ \\

 & Image-space flow & 0.981$_{\,(0.002)}$ & 0.878$_{\,(0.006)}$ & 0.936$_{\,(0.004)}$ & 0.152$_{\,(0.099)}$ & \textbf{0.011}$_{\,(0.016)}$ & 0.036$_{\,(0.044)}$ & 1.01$_{\,(0.39)}$ \\

 & RadCF & 0.986$_{\,(0.002)}$ & 0.827$_{\,(0.007)}$ & 0.908$_{\,(0.005)}$ & \textbf{0.094}$_{\,(0.073)}$ & 0.016$_{\,(0.028)}$ & 0.047$_{\,(0.097)}$ & \textbf{0.68}$_{\,(0.31)}$ \\
\bottomrule
\end{tabular}
}
\end{table*}

\begin{figure*}[t]
\centering
\includegraphics[width=\textwidth]{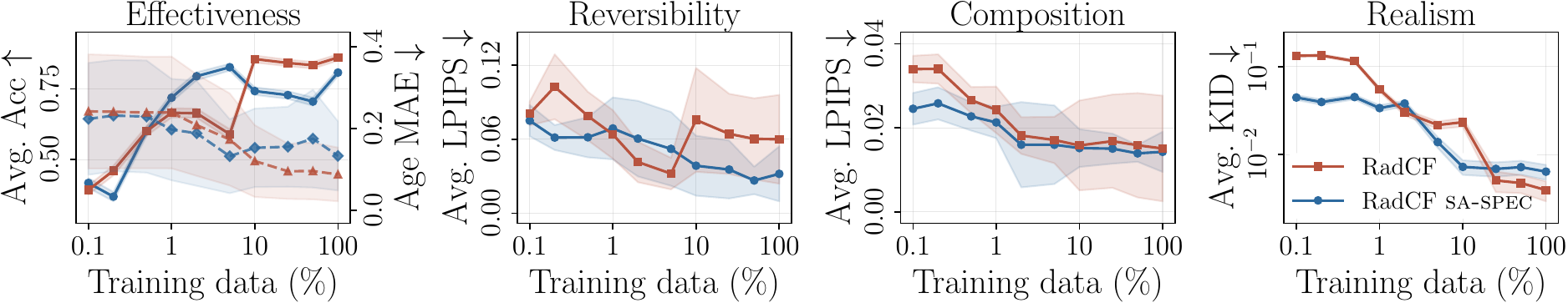}
\caption{\textbf{Specialisation is data-efficient.}
Trained on varying fractions of NIH-14 (1000 test samples). Solid lines average over the three categorical interventions; age is dashed and reported as MAE.
Per-intervention results in \Cref{appendix:data_efficiency_tables}.}
\label{fig:data_efficiency}
\end{figure*}

\noindent\textbf{Data and parameter efficiency.}
\Cref{fig:data_efficiency} plots counterfactual soundness as a function of target-domain data fraction (0.1\%--100\%, log scale), averaged across four interventions.
Specialisation matches the from-scratch baseline using only 24M LoRA parameters on the flow matching component (vs.\ 766M from scratch); \spec{sa-spec} additionally adapts the 84M-parameter VAE.
\spec{sa-spec} achieves the full-data view flip rate at only 5\% of NIH-14 ($0.993$ vs.\ $0.998$, while training from scratch at 5\% reaches only $0.684$; \Cref{appendix:data_efficiency_tables}).
Empirically, specialisation improves composition and reversibility in low-data regimes.
These results are consistent with \Cref{app:info_theory}: source pretraining reduces the marginal image-modelling burden, while specialisation learns the residual dependence on \(\pa\).
At 100\% data, specialisation retains a substantial composition and reversibility advantage at a modest cost to effectiveness on some interventions, a trade-off also observed by \citet{rasal2025diffusion}.

\begin{figure*}[t]
\centering
\begin{minipage}[t]{0.50\textwidth}
\centering
\includegraphics[width=0.63\linewidth]{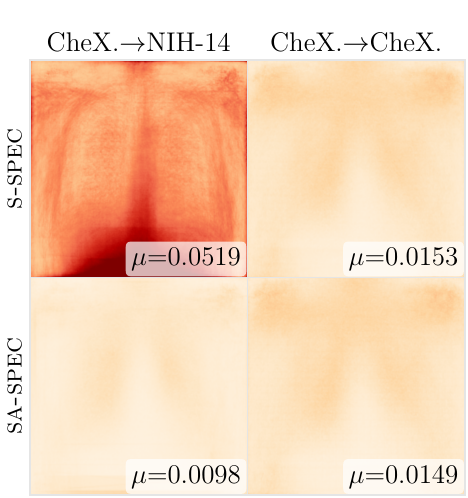}
\caption{
\textbf{\spec{sa-spec} reduces acquisition-shift distortion.}
Composition residuals for \spec{s-spec} and \spec{sa-spec} when a CheXpert-pretrained model is specialised to NIH-14 (out-of-domain) vs in-domain.
}
\label{fig:null_residuals}
\end{minipage}%
\hfill
\begin{minipage}[t]{0.48\textwidth}
\centering
\includegraphics[width=0.96\linewidth]{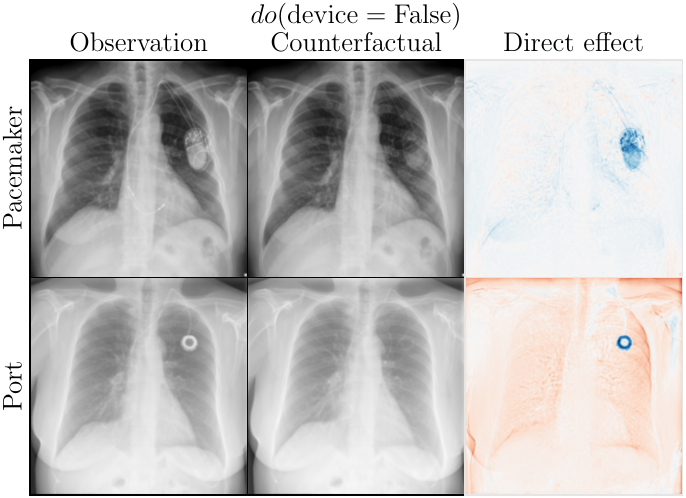}
\caption{
\textbf{\spec{sa-spec} enables device-removal counterfactuals on BRAX.}
Counterfactuals remove support devices under \(\doo(d)\)
while preserving subject-specific anatomy.
}
\label{fig:brax_device}
\end{minipage}
\end{figure*}

\noindent\textbf{When is \spec{sa-spec} necessary?} \Cref{tab:vae_isolation} isolates the role of acquisition specialisation by comparing \spec{s-spec} (flow-only) and \spec{sa-spec} (flow + VAE) under CheXpert\(\rightarrow\)NIH-14 acquisition shift and in-domain CheXpert\(\rightarrow\)CheXpert adaptation.
Under acquisition shift, \spec{sa-spec} substantially improves composition and reversibility while preserving effectiveness for most attributes.
In-domain, the same adaptation yields minimal change in composition and reversibility.
Full fine-tuning of the flow alone (\spec{s-spec}\textsubscript{full}) does not recover the same gains, indicating that the improvement comes from acquisition specialisation rather than from additional trainable parameters.
\Cref{fig:null_residuals} visualises this: under acquisition shift, \spec{s-spec} produces large pixel residuals around the lungs even on null interventions, indicating systematic decoder distortion of target-domain anatomy.
\spec{sa-spec} largely eliminates these residuals, while in-domain both methods produce comparably small residuals.

\begin{table*}[t]
\centering
\caption{\textbf{\spec{sa-spec} analysis: \spec{sa-spec} compensates for acquisition shift, improving composition and reversibility on the target domain (CheXpert$\to$NIH-14) but not on the source domain.}
Each effectiveness column reports pseudo-oracle performance on the intervened attributes; reversibility and realism are means across all four interventions.
KID is scaled by $\times 10^{2}$.
Best per column metrics are in bold. }
\label{tab:vae_isolation}
\small
\setlength{\tabcolsep}{4pt}
\resizebox{\textwidth}{!}{%
\begin{tabular}{ll cccc c c c}
\toprule
 & & \multicolumn{4}{c}{Effectiveness} & Composition & Reversibility & Realism \\
\cmidrule(lr){3-6}
\cmidrule(lr){7-7}
\cmidrule(lr){8-8}
\cmidrule(lr){9-9}
Domain & Method & $\doo(v)\!\uparrow$ & $\doo(e)\!\uparrow$ & $\doo(s)\!\uparrow$ & $\doo(a)\!\downarrow$ & LPIPS\,$\downarrow$ & LPIPS\,$\downarrow$ & KID\,$\downarrow$ \\
\midrule
 \multirow{3}{*}{\shortstack{CheXpert\\$\to$NIH-14}} & \spec{s-spec} & 0.976$_{\,(0.003)}$ & 0.359$_{\,(0.009)}$ & 0.884$_{\,(0.006)}$ & 0.170$_{\,(0.174)}$ & 0.092$_{\,(0.043)}$ & 0.163$_{\,(0.054)}$ & 3.13$_{\,(0.27)}$ \\
  & \spec{sa-spec} & \textbf{0.992}$_{\,(0.002)}$ & \textbf{0.565}$_{\,(0.009)}$ & 0.874$_{\,(0.006)}$ & 0.116$_{\,(0.079)}$ & \textbf{0.015}$_{\,(0.007)}$ & \textbf{0.034}$_{\,(0.021)}$ & \textbf{0.97}$_{\,(0.15)}$ \\
  & \spec{s-spec}\textsubscript{full} & 0.971$_{\,(0.003)}$ & 0.549$_{\,(0.009)}$ & \textbf{0.944}$_{\,(0.004)}$ & \textbf{0.111}$_{\,(0.081)}$ & 0.106$_{\,(0.045)}$ & 0.243$_{\,(0.057)}$ & 1.83$_{\,(0.21)}$ \\
\midrule
 \multirow{2}{*}{\shortstack{CheXpert\\$\to$CheXpert}} & \spec{s-spec} & \textbf{0.965}$_{\,(0.003)}$ & \textbf{0.478}$_{\,(0.010)}$ & 0.657$_{\,(0.007)}$ & 0.201$_{\,(0.139)}$ & \textbf{0.024}$_{\,(0.003)}$ & 0.037$_{\,(0.009)}$ & 1.34$_{\,(0.19)}$ \\
  & \spec{sa-spec} & 0.925$_{\,(0.004)}$ & 0.467$_{\,(0.010)}$ & \textbf{0.688}$_{\,(0.007)}$ & \textbf{0.166}$_{\,(0.102)}$ & 0.025$_{\,(0.004)}$ & \textbf{0.027}$_{\,(0.006)}$ & \textbf{1.01}$_{\,(0.19)}$ \\
\bottomrule
\end{tabular}
}
\end{table*}

\noindent\textbf{Extending $\pa$ in a new target domain.}
We apply \spec{sa-spec} to BRAX \citep{brax}, adding device presence to the target-domain $\pa$ to enable $\doo(d)$ counterfactuals.
Device is a localised attribute correlated with view position and pathology, and BRAX differs distinctly from the source domain in acquisition and population.
Despite this, composition and reversibility remain comparable to NIH-14 ($0.018$ and $0.019$ LPIPS, against $0.015$ and $0.034$).
Device interventions achieve a target accuracy of 0.729, and \Cref{fig:brax_device} shows device removal preserving patient-specific anatomy.
Per-intervention results are in \Cref{appendix:brax}.

\begin{figure*}[t]
\centering
\includegraphics[width=\textwidth]{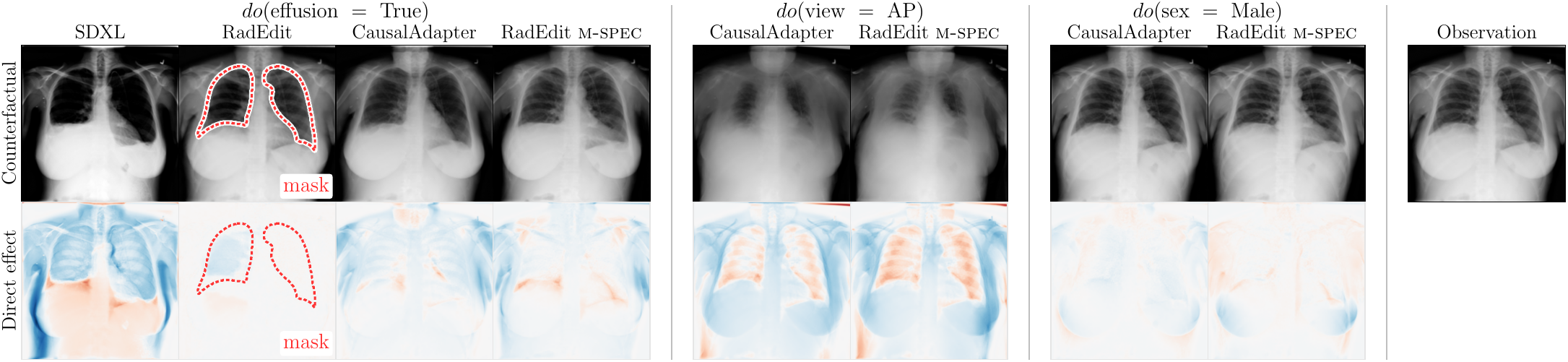}
\caption{
\textbf{Both structured specialisation methods generate sound counterfactuals without the need for masks.}
Counterfactuals (top) and direct-effect maps (bottom) are shown for representative disease, view, and sex interventions.
Compared with Causal-Adapter, RadEdit \spec{m-spec} more reliably changes the intended attribute while limiting collateral changes to patient-specific anatomy.
For disease editing, mask-guided RadEdit localises the intervention but requires a mask, whereas text-only SDXL introduces broader anatomical changes.
}
\label{fig:radedit_panel}
\end{figure*}

\subsection{Causal text-to-image generators with \spec{m-spec}}
\label{sec:result_mspec}

\Cref{tab:radedit_results} reports counterfactual soundness on 1{,}000 NIH-14 test samples.
We compare our \spec{m-spec} version of RadEdit against three methods: Causal-Adapter~\citep{tong2026causaladapter}, SDXL, and mask-guided RadEdit for disease editing.
For a controlled comparison, Causal-Adapter is implemented on the same RadEdit backbone as \spec{m-spec}; results on the authors' original Stable Diffusion 1.5 backbone are in \Cref{app:causal_adapter}.

The result first demonstrates the benefit of structured specialisation.
Under our taxonomy, both RadEdit \spec{m-spec} and Causal-Adapter~\citep{tong2026causaladapter} are instances of \spec{m-spec}: they both turn text-conditioned models into intervened causal variable conditions in \(\pa\).
For \(\doo(e)\), both \spec{m-spec} instances preserve non-target attributes better than text-only SDXL without requiring a localisation mask, indicating more targeted counterfactual changes.
RadEdit \spec{m-spec} also improves effectiveness over RadEdit.
This broader intervention interface comes with a trade-off: mask-guided RadEdit retains better composition, reversibility, and realism, as shown in \Cref{fig:radedit_panel}.

We next isolate the choice of specialisation mechanism.
On the same RadEdit backbone, \spec{m-spec} achieves better target-control point estimates for all four interventions and consistently lower composition error, reversibility error, and KID than Causal-Adapter.
We hypothesise that this advantage arises from the more direct integration of structured variables into RadEdit's pretrained conditioning pathway: \spec{m-spec} replaces text embeddings with metadata embeddings at the native conditioning interface, whereas
Causal-Adapter first aligns causal attributes with textual tokens and injects their interactions through an additional adapter.
This simpler, backbone-aligned parameterisation may make the structured conditions easier to optimise while better preserving the pretrained image distribution.

\begin{table*}[t]
\centering
\caption{
\textbf{Structured specialisation extends RadEdit to mask-free interventions on all variables in $\pa$, with RadEdit \spec{m-spec} outperforming Causal-Adapter on the same backbone.}
Relative to Causal-Adapter, RadEdit \spec{m-spec} achieves better target-control point estimates for every intervention and consistently lower composition error, reversibility error, and KID.
SDXL and mask-guided RadEdit are included only for $\doo(e)$; RadEdit additionally requires a localisation mask.
KID is scaled by $\times 10^{2}$.
Best per column metrics are in bold. }
\label{tab:radedit_results}
\small
\setlength{\tabcolsep}{4pt}
\resizebox{\textwidth}{!}{%
\begin{tabular}{ll cccc c c c}
\toprule
 & & \multicolumn{4}{c}{Effectiveness} & Composition & Reversibility & Realism \\
\cmidrule(lr){3-6}
\cmidrule(lr){7-7}
\cmidrule(lr){8-8}
\cmidrule(lr){9-9}
Intervention & Method & $\text{Acc}(v)\uparrow$ & $\text{Acc}(e)\uparrow$ & $\text{Acc}(s)\uparrow$ & $\text{MAE}(a)\downarrow$ & LPIPS\,$\downarrow$ & LPIPS\,$\downarrow$ & KID\,$\downarrow$ \\
\midrule
 \multirow{2}{*}{$\doo(v)$}
 & RadEdit Causal-Adapter & 0.563$_{\,(0.031)}$ & 0.658$_{\,(0.029)}$ & \textbf{0.963}$_{\,(0.012)}$ & 0.073$_{\,(0.061)}$ & 0.094$_{\,(0.072)}$ & 0.278$_{\,(0.091)}$ & 4.64$_{\,(0.61)}$ \\

 & RadEdit \spec{m-spec} & \textbf{0.600}$_{\,(0.030)}$ & \textbf{0.670}$_{\,(0.029)}$ & 0.947$_{\,(0.014)}$ & \textbf{0.067}$_{\,(0.056)}$ & \textbf{0.061}$_{\,(0.057)}$ & \textbf{0.202}$_{\,(0.067)}$ & \textbf{2.69}$_{\,(0.50)}$ \\
\midrule
 \multirow{4}{*}{$\doo(e)$}
 & SDXL & 0.748$_{\,(0.027)}$ & \textbf{0.776}$_{\,(0.027)}$ & 0.680$_{\,(0.029)}$ & 0.150$_{\,(0.115)}$ & \textbf{0.015}$_{\,(0.003)}$ & 0.488$_{\,(0.099)}$ & 3.54$_{\,(0.39)}$ \\

 & RadEdit & 0.995$_{\,(0.004)}$ & 0.566$_{\,(0.031)}$ & 0.972$_{\,(0.010)}$ & 0.070$_{\,(0.056)}$ & 0.027$_{\,(0.015)}$ & \textbf{0.107}$_{\,(0.050)}$ & \textbf{1.08}$_{\,(0.24)}$ \\

 & RadEdit Causal-Adapter & 0.979$_{\,(0.009)}$ & 0.531$_{\,(0.032)}$ & 0.974$_{\,(0.010)}$ & 0.065$_{\,(0.058)}$ & 0.094$_{\,(0.072)}$ & 0.223$_{\,(0.108)}$ & 1.85$_{\,(0.42)}$ \\

 & RadEdit \spec{m-spec} & \textbf{1.000}$_{\,(0.000)}$ & 0.682$_{\,(0.030)}$ & \textbf{0.976}$_{\,(0.009)}$ & \textbf{0.053}$_{\,(0.043)}$ & 0.061$_{\,(0.057)}$ & 0.174$_{\,(0.070)}$ & 1.40$_{\,(0.39)}$ \\
\midrule
 \multirow{2}{*}{$\doo(s)$}
 & RadEdit Causal-Adapter & 0.986$_{\,(0.007)}$ & 0.768$_{\,(0.026)}$ & 0.160$_{\,(0.023)}$ & 0.063$_{\,(0.054)}$ & 0.094$_{\,(0.072)}$ & 0.186$_{\,(0.106)}$ & 1.64$_{\,(0.45)}$ \\

 & RadEdit \spec{m-spec} & \textbf{0.997}$_{\,(0.003)}$ & \textbf{0.839}$_{\,(0.023)}$ & \textbf{0.547}$_{\,(0.031)}$ & \textbf{0.048}$_{\,(0.039)}$ & \textbf{0.061}$_{\,(0.057)}$ & \textbf{0.150}$_{\,(0.070)}$ & \textbf{0.75}$_{\,(0.26)}$ \\
\midrule
 \multirow{2}{*}{$\doo(a)$}
 & RadEdit Causal-Adapter & 0.981$_{\,(0.008)}$ & 0.790$_{\,(0.025)}$ & \textbf{0.976}$_{\,(0.009)}$ & 0.200$_{\,(0.133)}$ & 0.094$_{\,(0.071)}$ & 0.213$_{\,(0.107)}$ & 1.67$_{\,(0.39)}$ \\

 & RadEdit \spec{m-spec} & \textbf{0.993}$_{\,(0.005)}$ & \textbf{0.848}$_{\,(0.022)}$ & 0.962$_{\,(0.012)}$ & \textbf{0.191}$_{\,(0.126)}$ & \textbf{0.062}$_{\,(0.058)}$ & \textbf{0.143}$_{\,(0.077)}$ & \textbf{0.57}$_{\,(0.22)}$ \\
\bottomrule
\end{tabular}
}
\end{table*}

\subsection{Practicality and utility}
\label{sec:result_utility}

While our approach improves on current deep generative SCM methods under data scarcity and domain shift, counterfactual inference requires strong assumptions.
In particular, causal sufficiency is assumed: no unobserved confounders are taken into account, and only observed parents can be controlled for or intervened upon.
This assumption underpins most, if not all, deep-SCM counterfactual image generation \citep{pawlowski2020deepscm,monteiro2023axiomatic,ribeiro2023high,rasal2025diffusion,ribeiro2025counterfactual}.
Despite a likely violation in real data, counterfactual images may still be practically useful.
We support this claim with additional analyses of their downstream utility, since the assumption cannot be tested directly from observational data \citep{pan2024counterfactual}.

\noindent\textbf{Exposing shortcut learning.}
Counterfactuals reveal whether a diagnostic model relies on biased signals in the data to predict the disease status of a patient, i.e.\ shortcut reliance \citep{geirhos2020}.
When the bias is also present in the test set, standard accuracy metrics cannot detect shortcut learning, which is revealed only on an unbiased test set \citep{brown2023} that is not always available; intervening on the suspect variable removes that requirement.
Simulating ten levels of spurious view--effusion correlation, the rate at which the effusion prediction changes under $\doo(v)$ rises from $11\%$ at no bias to $47\%$ at the highest, monotonically across bias levels (Spearman's $\rho = 0.94$, $p < 0.05$), while AUC on real images stays within $0.72$-$0.84$ across all ten models (\Cref{appendix:utilityExperiments}).
Counterfactuals therefore detect shortcut learning that standard accuracy metrics miss in the absence of an unbiased test set.

\noindent\textbf{Debiasing via counterfactual augmentation.}
In the same setting, augmenting the biased training data with \spec{sa-spec} view-flipped counterfactuals reduces the reliance it exposes.
Against minority upsampling, which duplicates under-represented view--effusion combinations, and standard augmentation with cropping, rotation, colour jitter and erasing, only counterfactual augmentation holds shortcut reliance near its unbiased level across all bias settings, at a change rate of $0.115$ under the strongest bias against $0.593$ untreated (\Cref{appendix:utilityExperiments}).

\noindent\textbf{Composition with an upstream SCM.}
Previous sections assess the effect of changing one parent at a time while holding the remaining parents at their factual values.
The specialisation mechanism is however not restricted to that setting and accepts any parent value a causal graph supplies.
To show this, we incorporate our image mechanism into a deep SCM defined as $a \rightarrow e \rightarrow v$, $\{s,a,e,v\} \rightarrow \x$.
The edge $e \rightarrow v$ encodes a plausible acquisition mechanism: patients with more severe effusion are less mobile and are therefore more often imaged supine (AP) rather than standing (PA).

We compare a view-fixed edit with an SCM-guided $\doo(e)$ query that allows the intervention to propagate through $e \rightarrow v$.
In the cases where the SCM prescribed a view change, our method reaches the same effusion effectiveness ($0.827$) under both edits; while SCM-guided edits incur higher reversibility error
and lower fidelity than view-fixed edits.
Detailed results can be found in \Cref{appendix:scm_propagation}.

\section{Discussion and limitations}
\label{sec:conclusion}

In this work, we show that pretrained generative models can be \emph{specialised} into image mechanisms for counterfactual generation.
Our central contribution is to frame counterfactual image generation as an adaptation problem: reusing pretrained image knowledge to learn a target mechanism rather than training from scratch.
This perspective complements adapter-based approaches such as Causal-Adapter \citep{tong2026causaladapter}.

Our study examines adaptation under limited data, distribution shift, and changing intervention requirements, addressing practical challenges in medical imaging.
RadCF implements this framework through latent flow matching, providing competitive counterfactual soundness and image fidelity on chest X-rays.
Specialisation also applies beyond flow-based mechanisms.
In particular, \spec{m-spec} adapts a text-conditional diffusion model to enable structured interventions, without needing text prompts or auxiliary spatial annotations.
In doing so, we take a step toward using foundation models for counterfactual reasoning, with applications in model auditing and data augmentation.


\noindent\textbf{Limitations and future work.}
While our experiments target measurement bias and acquisition shift, specialised mechanisms may still be affected by manifestation shift, where the same clinical factor induces different image-level appearances across populations.
Evaluating zero-shot transfer under such shifts is an important direction for future work.
To scale to high-resolution counterfactual images, RadCF parameterises the mechanism in latent space.
This differs from the exactly invertible image-space mechanisms studied in flow-based counterfactual identification \citep{ribeiro2025counterfactual}.
Our mechanism also follows the causal Markovian assumption.
This assumption may be violated in clinical data by unobserved confounding, but we provide \Cref{appendix:probe} which examines how closely this assumption is satisfied in our datasets.
Extending specialisation to non-Markovian settings remains future work.

\bibliographystyle{plainnat}
\bibliography{scmadapter}

\newpage
\appendix
\crefalias{section}{appendix}
\crefalias{subsection}{appendix}
\section{Information-theoretic motivation for specialisation}
\label{app:info_theory}
We consider the setting where a generative model is trained on a source domain \(p_s(\x)\), or on a source text-conditional distribution \(p_s(\x\mid\spec{TEXT})\), and then specialised on a target domain \(p_t(\x,\pa)\).
In our experiments, the source domain corresponds to CheXpert or a pretrained radiology foundation model, while the target domain corresponds to NIH-14, BRAX, or a structured-metadata counterfactual task.

Throughout this appendix we analyse an idealised adaptation gap. We assume that, in the target domain, the specified image mechanism is causally sufficient for
\(\x\), i.e.
\[
    \x = f_t(\uu,\pa), \qquad \uu \indep \pa .
\]
Under this assumption,
\[
    p_t(\x\mid \doo(\pa)) = p_t(\x\mid \pa).
\]
The following results should be interpreted as distributional motivation for specialisation, not as finite-sample generalisation guarantees.

\paragraph{Proposition 1: Source pretraining decomposes target adaptation into structure and acquisition mismatch.}

Assume the source-pretrained unconditional generator represents the source image marginal \(p_s(\x)\).
Define the target adaptation gap before introducing explicit parent conditioning as
\[
    \mathcal{G}_{\spec{S}}(p_s)
    :=
    \mathbb{E}_{p_t(\pa)}
    \left[
        \operatorname{KL}
        \left(
            p_t(\x\mid \doo(\pa))
            \,\|\, p_s(\x)
        \right)
    \right].
\]
Then
\[
    \mathcal{G}_{\spec{S}}(p_s)
    =
    I_t(\x;\pa)
    +
    \operatorname{KL}
    \left(
        p_t(\x)\,\|\,p_s(\x)
    \right),
\]
where \(I_t(\x;\pa)\) denotes mutual information under the target distribution
\(p_t(\x,\pa)\).

\paragraph{Proof.}

Using \(p_t(\x\mid \doo(\pa))=p_t(\x\mid\pa)\), we have
\[
\begin{aligned}
\mathcal{G}_{\spec{S}}(p_s)
&=
\mathbb{E}_{p_t(\pa)}
\mathbb{E}_{p_t(\x\mid\pa)}
\left[
    \log p_t(\x\mid\pa) - \log p_s(\x)
\right] \\
&=
\mathbb{E}_{p_t(\pa)}
\mathbb{E}_{p_t(\x\mid\pa)}
\left[
    \log p_t(\x\mid\pa) - \log p_t(\x)
\right] \\
&\quad+
\mathbb{E}_{p_t(\pa)}
\mathbb{E}_{p_t(\x\mid\pa)}
\left[
    \log p_t(\x) - \log p_s(\x)
\right] \\
&=
I_t(\x;\pa)
+
\operatorname{KL}
\left(
    p_t(\x)\,\|\,p_s(\x)
\right).
\end{aligned}
\]
The last equality follows from the definition of mutual information and the tower property
\(\mathbb{E}_{p_t(\pa)}\mathbb{E}_{p_t(\x\mid\pa)}[\cdot]
=
\mathbb{E}_{p_t(\x)}[\cdot]\).

\paragraph{Interpretation.}

Proposition 1 separates the target adaptation gap into two terms.
The first term, \(I_t(\x;\pa)\), is the parent-dependent structure needed for counterfactual control.
This is the term targeted by \spec{s-spec}, which introduces explicit conditioning on \(\pa\) through low-rank updates to the pretrained mechanism.

The second term,
\[
    \operatorname{KL}\left(p_t(\x)\,\|\,p_s(\x)\right),
\]
is the source-target marginal mismatch.
In our setting, this corresponds to acquisition shift, scanner differences, image preprocessing differences, and population differences between source and target domains.
This term motivates \spec{sa-spec}: adapting the autoencoder helps align the image marginal of the source-pretrained model with the target domain, while the flow learns the parent-dependent structure.
Thus, when \(p_s(\x)\) is already close to \(p_t(\x)\), \spec{s-spec} may be sufficient; when the marginal mismatch is large,
\spec{sa-spec} is expected to help.

\paragraph{Proposition 2: Text-conditional pretraining reduces the residual structural gap.}

We now consider a source model pretrained on \(p_s(\x\mid\spec{TEXT})\).
Since the target domain is specified by structured metadata \(p_t(\x,\pa)\), we introduce an idealised measurement model \(r(\spec{TEXT}\mid\pa)\), which represents the text proxy associated with \(\pa\).
This induces the joint distribution
\[
    p_t^r(\x,\pa,\spec{TEXT})
    :=
    p_t(\x,\pa)\,r(\spec{TEXT}\mid\pa).
\]
By construction, this idealised proxy satisfies
\[
    \x \indep \spec{TEXT} \mid \pa .
\]

Define the target adaptation gap for a source text-conditional model as
\[
    \mathcal{G}_{\spec{M}}(p_s)
    :=
    \mathbb{E}_{p_t^r(\pa,\spec{TEXT})}
    \left[
        \operatorname{KL}
        \left(
            p_t(\x\mid \doo(\pa))
            \,\|\, p_s(\x\mid\spec{TEXT})
        \right)
    \right].
\]
Then
\[
    \mathcal{G}_{\spec{M}}(p_s)
    =
    I_{t,r}(\x;\pa\mid\spec{TEXT})
    +
    \mathbb{E}_{p_t^r(\spec{TEXT})}
    \operatorname{KL}
    \left(
        p_t^r(\x\mid\spec{TEXT})
        \,\|\, p_s(\x\mid\spec{TEXT})
    \right),
\]
where \(I_{t,r}\) denotes mutual information under
\(p_t^r(\x,\pa,\spec{TEXT})\).

\paragraph{Proof.}

Using \(p_t(\x\mid \doo(\pa))=p_t(\x\mid\pa)\), we expand
\[
\begin{aligned}
\mathcal{G}_{\spec{M}}(p_s)
&=
\mathbb{E}_{p_t^r(\pa,\spec{TEXT})}
\mathbb{E}_{p_t(\x\mid\pa)}
\left[
    \log p_t(\x\mid\pa)
    -
    \log p_s(\x\mid\spec{TEXT})
\right] \\
&=
\mathbb{E}_{p_t^r(\pa,\spec{TEXT})}
\mathbb{E}_{p_t(\x\mid\pa)}
\left[
    \log p_t(\x\mid\pa)
    -
    \log p_t^r(\x\mid\spec{TEXT})
\right] \\
&\quad+
\mathbb{E}_{p_t^r(\pa,\spec{TEXT})}
\mathbb{E}_{p_t(\x\mid\pa)}
\left[
    \log p_t^r(\x\mid\spec{TEXT})
    -
    \log p_s(\x\mid\spec{TEXT})
\right].
\end{aligned}
\]
Because \(p_t^r(\x,\pa,\spec{TEXT})
=
p_t(\x,\pa)r(\spec{TEXT}\mid\pa)\), we have
\(\x \indep \spec{TEXT}\mid\pa\). Therefore, the first term is
\[
    H_{t,r}(\x\mid\spec{TEXT}) - H_t(\x\mid\pa)
    =
    I_{t,r}(\x;\pa\mid\spec{TEXT}).
\]
For the second term, marginalising over \(\pa\) gives
\[
    \mathbb{E}_{p_t^r(\spec{TEXT})}
    \operatorname{KL}
    \left(
        p_t^r(\x\mid\spec{TEXT})
        \,\|\,p_s(\x\mid\spec{TEXT})
    \right).
\]
Combining these terms proves the result.

\paragraph{Comparison with unconditional source pretraining.}

Let
\[
    \delta_{\spec{S}}
    =
    \operatorname{KL}\left(p_t(\x)\,\|\,p_s(\x)\right)
\]
denote the unconditional source-target marginal mismatch, and let
\[
    \delta_{\spec{M}}
    =
    \mathbb{E}_{p_t^r(\spec{TEXT})}
    \operatorname{KL}
    \left(
        p_t^r(\x\mid\spec{TEXT})
        \,\|\,p_s(\x\mid\spec{TEXT})
    \right)
\]
denote the text-conditional source-target mismatch. Combining Propositions 1
and 2 gives
\[
    \mathcal{G}_{\spec{S}}(p_s)
    -
    \mathcal{G}_{\spec{M}}(p_s)
    =
    I_{t,r}(\x;\spec{TEXT})
    +
    \delta_{\spec{S}}
    -
    \delta_{\spec{M}}.
\]
Thus, text-conditional source pretraining reduces the idealised adaptation gap whenever the information about the target image contained in the text proxy, \(I_{t,r}(\x;\spec{TEXT})\), together with the unconditional source-target alignment, outweighs the conditional source-target mismatch.

\paragraph{Interpretation.}

Proposition 2 motivates \spec{m-spec}.
Text-conditional radiology models have already learned image variations associated with textual descriptions. If \(\spec{TEXT}\) is informative about \(\pa\), the remaining structural gap is reduced from \(I_t(\x;\pa)\) to \(I_{t,r}(\x;\pa\mid\spec{TEXT})\), up to conditional source-target mismatch.
\spec{m-spec} reuses this text-conditional image prior while replacing imprecise text prompts with structured metadata embeddings \(g_\omega(\pa)\).
This gives explicit intervention control over \(\pa\) while preserving semantic knowledge from the pretrained text-conditional backbone.

The assumption \(\x \indep \spec{TEXT}\mid\pa\) is idealised.
Radiology text can contain information beyond the selected variables in \(\pa\), and structured metadata can itself be noisy. Therefore, Proposition 2 should be interpreted as an adaptation-gap motivation for measurement specialisation, rather than a guarantee that text conditioning is sufficient for counterfactual control.

\section{Related work}
\label{app:related_work}
\paragraph{Counterfactual image generation.}
Counterfactual image generation aims to answer retrospective questions about how an observed instance would have appeared under an alternative intervention, while preserving all non-descendant, instance-specific factors.
This is naturally formalised using structural causal models (SCMs), where counterfactuals are computed through abduction, action, and prediction \citep{pearl2009causality,peters2017elements,Scholkopfetal21}.
Early deep SCMs parameterised image mechanisms with deep generative models, enabling counterfactual generation in settings where the image is treated as a high-dimensional endogenous variable \citep{pawlowski2020deepscm}.
Since ground-truth counterfactual images are generally unavailable, subsequent work has evaluated learned mechanisms through axiomatic criteria such as composition, reversibility, and effectiveness \citep{galles1998axiomatic,halpern2000axiomatizing,monteiro2023axiomatic}, with realism later incorporated as an additional fidelity criterion \citep{melistas2025benchmarking}.
These metrics are now widely used to quantify whether generated counterfactuals preserve identity, obey interventions, and remain on the image manifold.

Building on these, hierarchical VAEs \citep[HVAEs, ][]{vahdat2020nvae, child2020very}, diffusion models \citep{song2021denoising, preechakul2022diffusion, rombach2022high}, and optimal-transport flow matching \citep{lipman2023flow, pmlr-v202-pooladian23a, tong2024improving} have been used to parameterise mechanisms for generating high-fidelity counterfactuals in mammography, 2D and 3D brain MRIs, subcortical meshes and chest X-rays \citep{ribeiro2023high, ribeiro2025counterfactual, rasal2025diffusion, rasal2022deep, xia2025decoupledcfg, peng2025latent}.
However, most existing counterfactual image generation methods train mechanisms from scratch on the target domain.
This limits their applicability in clinical settings, where annotated data are scarce, costly to obtain, and difficult to pool across institutions because of privacy and governance constraints \citep{willemink2020preparing,kaissis2020secure}.
In contrast, our work adapts pretrained generative models into causal mechanisms, reducing the need for large target-domain datasets.

\paragraph{Medical image generation and editing with foundation models.}
General-purpose image editing methods such as DiffEdit and text-guided editing localise changes through masks or textual instructions \citep{couairon2023diffedit,lin2024text}, and analogous approaches have been adopted in medical imaging.
In radiology, recent models have used text, reports, masks, or other conditioning signals to synthesise or modify chest X-rays with improved visual fidelity and controllability \citep{Bluethgen2024,weber2023cascaded,pinaya2023generative,trang2025discovering,gu2023biomedjourney,perez-garcia_bond-taylor_radedit,kumar2025prism,cooke2025roentmod}.
These models provide strong image priors and can perform clinically meaningful edits, but their control is usually associative rather than counterfactual: prompts, masks, or bounding boxes specify desired visual changes without explicitly modelling the causal mechanisms that generated the observation.
However, mask-based methods require spatial annotations that are often unavailable in clinical datasets \citep{xing2023less,wang2021annotation,ma2024segment}, and local masks may be insufficient for global interventions involving acquisition view, demographic attributes, or anatomical morphology.
Prompt-based control also introduces ambiguity, since clinically relevant causal variables may be only indirectly represented in free text.

Our work differs from medical image editing in that we specialise pretrained image models to parameterise explicit SCM mechanisms.
Rather than asking a model to edit an image through a prompt or mask, we abduct an exogenous code from the observed image and regenerate the image under intervened parent variables.
This connects the flexibility of foundation image models with the formal semantics of counterfactual inference.

\paragraph{Domain shift and causal modelling in medical imaging.}
Medical imaging datasets often differ across hospitals, scanners, protocols, patient populations, and labelling pipelines.
Such acquisition shifts and measurement biases can degrade model performance and lead to unreliable conclusions when models are transferred across datasets \citep{Castro2020-ic,roschewitz2023automatic}.
Causal approaches have been proposed as a way to reason about these shifts, since SCMs separate stable mechanisms from changing marginal distributions and provide a language for interventions and counterfactuals \citep{Scholkopfetal21,bareinboim2022on,richens2020improving}.
In medical imaging, causal structure has been used to study confounding, improve robustness, and define clinically meaningful image interventions \citep{roschewitz2024counterfactual,XiaTia_Segmentorguided_MICCAI2025,MehRag_CFSeg_MICCAI2025}.
However, most work either assumes that the relevant mechanisms can be learned directly from the target domain, or focuses on associative and interventional robustness rather than full counterfactual image generation.
Our setting is more demanding: the target domain may contain limited labelled data, may differ from the source distribution used for pretraining, and may require interventions on variables that were not explicitly modelled during pretraining.

\paragraph{Parameter-efficient adaptation of generative models.}
Parameter-efficient fine-tuning has emerged as a practical way to adapt large pretrained models without updating all weights.
Low-Rank Adaptation (LoRA) learns low-rank weight updates while keeping the pretrained backbone fixed, reducing trainable parameters and improving fine-tuning efficiency \citep{hu2022lora}.
Adapter-based methods have also been used to add new forms of conditioning to pretrained image models \citep{zhang2023adding}.
In the causal setting, Causal-Adapter \citep{tong2026causaladapter} is closest to our measurement-specialisation setting: it adapts a frozen text-to-image diffusion backbone by learning causal mechanisms over semantic attributes and injecting them through adapter and token-level conditioning.
In contrast, our specialisation framework targets radiology counterfactual mechanisms under dataset shift: \spec{s-spec} adds explicit parent conditioning to pretrained generators, \spec{sa-spec} additionally adapts the latent representation and generator to acquisition-shifted target domains, and \spec{m-spec} replaces text conditioning with structured clinical parent variables.

\section{Experimental details}
\label{appendix:experiment_details}
\subsection{Datasets and preprocessing}
\label{appendix:datasets}

We use three publicly available chest radiograph datasets summarised in \Cref{tab:dataset_stats}.
All three are released under data use agreements for research purposes; no additional institutional review board approval was required.

\noindent\textbf{CheXpert.}
The base generator is trained on 223{,}414 CheXpert radiographs (191{,}027 frontal and 32{,}387 lateral images) from 64{,}540 patients at Stanford Hospital \citep{irvin2019chexpert}.
Patient age ranges from 0 to 90 years (mean $60.4\pm17.8$); 59.4\% are male; effusion prevalence is 43.8\%.
For the CheXpert$\to$CheXpert adaptation experiment, we filter to 33{,}454 frontal images from 13{,}020 patients (16{,}499 training, 16{,}955 evaluation).
Although these images were seen during base-model pretraining, the base model was not conditioned on the intervention variables; this setting tests whether \spec{s-spec} alone suffices when no acquisition shift is present.

\noindent\textbf{NIH-14.}
NIH ChestX-ray14 contains 112{,}120 frontal chest radiographs from 30{,}805 patients \citep{wang2017chestx}.
We discard 16 images with implausible ages ($>$100 years), yielding 112{,}104 images from 30{,}802 patients.
Pathology labels for 14 conditions, including effusion, were extracted by the original dataset authors using NLP applied to associated radiology reports; these are \emph{not} derived from
the CheXpert labeller.
We use 100{,}000 images for specialisation training and hold out the remaining 12{,}104 images for evaluation, with no patient-level overlap between the two sets.
NIH-14 differs from CheXpert in institution, scanner hardware, image resolution, and acquisition protocol, introducing the acquisition shift that motivates the VAE component of
\spec{sa-spec}.
Patient age ranges from 1 to 100 years (mean $46.9\pm16.6$); 56.5\% are male; 60.0\% are PA views; effusion prevalence is 11.9\%.

\noindent\textbf{BRAX.}
BRAX comprises 40{,}967 radiographs (AP, PA, and lateral) from 19{,}351 patients at a Brazilian institution \citep{brax}.
We retain only frontal views for consistency with the other datasets, yielding 19{,}309 images from 15{,}113 patients.
We observed substantial label noise in the BRAX device annotations, so we used a device classifier on Chexpert, and removed images whose device-classifier prediction confidence falls below 0.85, leaving 14{,}636 frontal images from 11{,}530 patients (11{,}636 training, 3{,}000 evaluation).
BRAX introduces a more extreme domain shift than NIH-14: patient demographics, disease prevalence, and imaging equipment differ substantially from CheXpert.
Patient age ranges from 0 to 85+ years (mean $45.7\pm26.5$); 61.6\% are male; 69.7\% are PA views; effusion prevalence is 5.4\%; device prevalence is 37.6\%.

\noindent\textbf{Intervention variables.}
Across all datasets, we extract four parent variables from structured metadata: view position $v \in \{\text{AP}, \text{PA}\}$, sex $s \in \{\text{M}, \text{F}\}$, age $a \in [0,1]$ (divided by 100), and effusion status $e \in \{0, 1\}$.
BRAX additionally provides a support device label $d \in \{0, 1\}$.
Binary labels are used as-is from the dataset metadata.

\begin{table}[t]
\centering
\small
\caption{Dataset summary statistics.}
\label{tab:dataset_stats}
\setlength{\tabcolsep}{3.5pt}
\begin{tabular}{lcccc}
\toprule
& CheXpert & CheXpert Frontal & NIH-14 & BRAX \\
\midrule
Total images           & 223{,}414 & 33{,}454 & 112{,}104 & 14{,}636 \\
~~Frontal / Lateral    & 191{,}027 / 32{,}387 & 33{,}454 / --- & 112{,}104 / --- & 14{,}636 / --- \\
Patients               & 64{,}540 & 13{,}020 & 30{,}802 & 11{,}530 \\
Finetune / Eval split  & 223{,}414 / --- & 16{,}499 / 16{,}955 & 100{,}000 / 12{,}104 & 11{,}636 / 3{,}000 \\
AP / PA (\%)           & 84.6 / 15.4 & 82.6 / 17.4 & 40.0 / 60.0 & 30.3 / 69.7 \\
Male (\%)              & 59.4 & 59.2 & 56.5 & 61.6 \\
Age (mean $\pm$ std)   & $60.4 \pm 17.8$ & $61.6 \pm 17.5$ & $46.9 \pm 16.6$ & $45.7 \pm 26.5$ \\
Effusion prevalence (\%) & 43.8 & 71.4 & 11.9 & 5.4 \\
Device prevalence (\%) & --- & --- & --- & 37.6 \\
\bottomrule
\end{tabular}
\end{table}

\noindent\textbf{Preprocessing.}
All images are resized to $256 \times 256$ pixels using direct resize with bilinear interpolation.
Images are converted to 3-channel RGB and pixel intensities are normalised to $[-1, 1]$.
Age is normalised by dividing by 100, mapping to approximately $[0, 1]$.
No histogram equalisation, windowing, or data augmentation is applied during specialisation.
For evaluation, counterfactual generation is performed on held-out patients with no overlap with the generative training set.

\subsection{Model architecture and training}
\label{appendix:model_architecture}
The base transport network is \texttt{SiT-XL/2} \citep{ma2024sit}.
Source-domain pretraining is performed end-to-end with the autoencoder and REPA-E alignment \citep{leng2025repae}.
The base model is trained for 400k steps on 4 GPUs with batch size 16.
Experiments were run on an internal HPC cluster using NVIDIA H100 (80GB) GPUs, scheduled via SLURM.

We represent the parent variables with a schema-driven embedding: categorical variables are mapped through learned embeddings, continuous variables are projected linearly. The resulting embeddings are summed into a single conditioning vector and injected through the transformer's adaLN-Zero conditioning pathway.

During $\spec{s-spec}$, LoRA is applied to attention, MLP, and conditioning projections, with rank $r\in\{4,8,16,32\}$ and corresponding scaling factors $\alpha\in\{8,16,32,64\}$, where $r = 32$ is chosen for the main results.
During $\spec{sa-spec}$, the autoencoder and transport are updated in alternating phases exactly as in \citet{leng2025repae}.
Counterfactual inference uses a 200-step Euler ODE solver for both inversion and generation; for the diffusion baseline, 1000 ODE steps are used; for HVAE, we use 20 latent layers.
During domain specialisation, LoRA modules are trained with AdamW ($\beta_1{=}0.9$, $\beta_2{=}0.999$, lr$\,{=}\,10^{-4}$) and batch size 16.
A class dropout probability of 0.1 is used during training; classifier-free guidance scale $w{=}1$ is applied at inference.

\subsection{Baseline implementations}
\label{appendix:baselines}

\noindent\textbf{HVAE} \citep{ribeiro2023high} is a hierarchical VAE operating at $256{\times}256$ with 8 resolution stages (35 encoder blocks, 42 decoder blocks of which 38 are stochastic at $4{\times}4$--$128{\times}128$, $z_\text{dim}{=}16$; 8.1M parameters).
Parent attributes are spatially broadcast and concatenated at every decoder block, giving a conditional prior $p(z_i \mid z_{<i}, \pa)$.
The model is trained from scratch on NIH-14 with AdamW (lr${=}10^{-3}$, wd${=}0.05$), optimising the negative ELBO with a discretised Gaussian likelihood.
The model is then fine-tuned with a Lagrangian objective combining a judge loss from five frozen ResNet-50 classifiers, an $\ell_1$ reconstruction penalty, and an ELBO constraint (AdamW, lr${=}10^{-4}$, batch size~8).

\noindent\textbf{Diffusion} \citep{rasal2025diffusion} adapts the spatial counterfactual mechanism of \citep{rasal2025diffusion} to a latent diffusion architecture.
We use a latent diffusion model based on the U-Net architecture of \citet{rombach2022high}, operating at $256{\times}256$ resolution with the same KL-regularised autoencoder as in RadCF.
Parent conditioning is injected via cross-attention.
The model uses the same training data splits as RadCF for fair comparison.
Counterfactual inference uses DDIM inversion \citep{song2021denoising} with 1000 steps for both the forward (abduction) and reverse (generation) passes.

\section{Evaluation implementation details}
\label{appendix:metricImplementations}

\subsection{Pseudo-oracle classifiers.}\label{appendix:oracles}

All classifiers implement $\x \rightarrow a$ and are trained exclusively on the training split of the corresponding evaluation dataset, ensuring no patient-level overlap with the generative evaluation data.
Training uses the Adam optimiser with an 80/20 train/validation split, checkpointing the epoch with best validation AUC for discrete attributes (view, sex, effusion, device) and lowest validation MAE for continuous attributes (age).
Binary decision thresholds are chosen on the validation set to maximise the sum of sensitivity and specificity, then applied to the test set.
Classifier performance is summarised in \Cref{tab:classifier_performance}.

\begin{table}[t]
\centering
\small
\begin{threeparttable}
\caption{Pseudo-oracle performance. Val.\ and Test report AUC for binary classifiers and MAE for the age regressor (normalised age $\in [0,1]$).}
\label{tab:classifier_performance}
\begin{tabular}{lllrcc}
\toprule
Dataset & Pseudo-oracle & Architecture & Best ep. & Val. & Test \\
\midrule
\multirow{5}{*}{NIH14}
    & View                     & ResNet-18    & 2  & 0.999 & 1.000 \\
    & Sex                      & ResNet-18    & 3  & 0.999 & 0.995 \\
    & Effusion (PA)\tnote{$\dagger$}   & DenseNet-121 & 9  & 0.881 & 0.863 \\
    & Effusion (AP)\tnote{$\dagger$}   & DenseNet-121 & 17 &  0.827 & 0.812 \\
    & Age                      & ResNet-18    & 9  & 0.041 & 0.051 \\
\midrule
\multirow{5}{*}{CheXpert}
    & View                     & ResNet-18    & 3  & 0.997 & 0.995 \\
    & Sex                      & ResNet-18    & 2  & 0.993 & 0.992 \\
    & Effusion (PA)\tnote{$\ddagger$}  & ResNet-18    & 13 & 0.935 & 0.911 \\
    & Effusion (AP)\tnote{$\ddagger$}  & ResNet-18    & 12 & 0.919 & 0.928 \\
    & Age                      & ResNet-18    & 9  & 0.067 & 0.072 \\
\midrule
\multirow{7}{*}{BRAX}
    & View                     & ResNet-18    & 2  & 0.999 & 0.999 \\
    & Sex                      & ResNet-18    & 2  & 0.984 & 0.982 \\
    & Effusion (PA)\tnote{$\S$} & ResNet-18    & 6  & 0.908 & 0.853 \\
    & Effusion (AP)\tnote{$\S$} & ResNet-18    & 11 & 0.807 & 0.663 \\
    & Age                      & ResNet-18    & 8  & 0.047 & 0.051 \\
    & Device (PA)              & ResNet-18    & 5  & 0.771 & 0.743 \\
    & Device (AP)              & ResNet-18    & 7  & 0.965 & 0.939 \\
\bottomrule
\end{tabular}
\begin{tablenotes}
    \item[$\dagger$] \citet{wang2017chestx} report Effusion AUC = 0.736 (8-class) and \citet{rajpurkar2017chexnet} report 0.864 (14-class), both training jointly across all labels on pooled frontal views without view stratification.
    \item[$\ddagger$] \citet{irvin2019chexpert} report Pleural Effusion AUC = 0.97 (multi-view, uncertainty-aware training, radiologist-annotated ground truth). Note that direct comparison is limited by multi-view input and the stronger reference standard.
    \item[$\S$] No published Effusion AUC benchmark exists for BRAX; values are reported for reference only.
\end{tablenotes}
\end{threeparttable}
\end{table}

\noindent\textbf{View and Sex.}
Both attributes are classified using a ResNet-18 (ImageNet pretrained) with a binary head.
Training ran for up to 3 epochs with lr$\,{=}\,10^{-4}$ and class-balanced minibatches, identically for NIH14, CheXpert, and BRAX.

\noindent\textbf{Effusion.}
Based on \citet{aslani2023optimising}, classifiers are trained separately on PA and AP subsets to remove the well-established systematic view-disease confound in chest radiographs \citep{raoof2012interpretation}.
All three datasets use label smoothing ($\varepsilon = 0.1$) and ReduceLROnPlateau (factor 0.2, patience 3 epochs).
On NIH14, ablation showed that domain-specific initialisation improves Effusion sensitivity: a DenseNet-121 is initialised from CheXpert-pretrained weights via TorchXRayVision \citep{pmlr-v172-cohen22a} and fine-tuned at lr$\,{=}\,10^{-5}$ for up to 20 epochs to preserve these representations.
On CheXpert, alternative domain-pretrained weights and the DenseNet-121 architecture were evaluated in ablation but offered no improvement, so an ImageNet-pretrained ResNet-18 is trained at lr$\,{=}\,10^{-4}$ for up to 15 epochs.
NIH14 and CheXpert classifiers are benchmarked against published Effusion classifiers in \Cref{tab:classifier_performance}, supporting their use as sufficiently performant pseudo-oracles; no published benchmark exists for BRAX.

At evaluation time, counterfactuals are routed to the PA or AP disease classifier by predicted view rather than original or target view.
This handles two edge cases: the model may accidentally flip view when intervening on another attribute, or a targeted view flip may fail.
In both cases, predicted-view routing ensures the disease classifier matches the image's actual acquisition geometry.
For AUC-based classifier calibration, selection uses ground truth view.
When view is the target intervention, counterfactual images are routed to the target-view disease classifier, measuring whether $\widetilde{\x}$ convinces that classifier.

\noindent\textbf{Age.}
A ResNet-18 with a sigmoid scalar output predicts normalised age in $[0,1]$, trained with Huber loss ($\delta = 0.1$) for robustness to label noise.
Training runs for up to 10 epochs at lr$\,{=}\,10^{-4}$ with ReduceLROnPlateau (factor = 0.2, patience = 2 epochs), identically for NIH14, CheXpert, and BRAX.
Direction success for age is computed only over samples where the requested change satisfies $|\widetilde{a} - a| > 10^{-3}$.
Smaller changes are excluded as trivial.

\noindent\textbf{Device (BRAX only).}
Two separate ResNet-18 classifiers (ImageNet-pretrained) with binary heads are trained, one per view (PA and AP), to account for view-dependent device appearance.
Training follows the same protocol as View and Sex: Adam optimiser at lr$\,{=}\,10^{-4}$, class-balanced minibatches, CrossEntropyLoss (no label smoothing), and ReduceLROnPlateau (factor 0.2, patience 3 epochs).
Decision thresholds are selected by Youden's J on the validation set.
At evaluation time, counterfactuals are routed to the PA or AP device classifier by predicted view, analogously to the Effusion routing described above.

\subsection{Evaluator sensitivity}\label{appendix:evaluator_sensitivity}

Effectiveness in the main text is measured with the pseudo-oracles of \Cref{appendix:oracles}.
While pseudo-oracles are widely adopted as a measure of effectiveness, we acknowledge the potential for evaluator bias to confound conclusions.
To investigate its impact on our results, we additionally re-scored every counterfactual with six independently trained public evaluators spanning different backbones, training paradigms and resolutions: four TorchXRayVision models~\citep{pmlr-v172-cohen22a}, CheXzero~\citep{tiu2022expert}, and a linear classifier fitted on frozen RAD-DINO features~\citep{perez-garcia2025raddino}.
Each evaluator is thresholded at its own Youden-optimal operating point on the NIH-14 train split rather than at $0.5$, so that flip rates are not confounded by differences in classifier calibration.
The evaluators themselves differ substantially in quality, spanning effusion AUC $0.756$ to $0.864$, allowing us to compare oracle-based effectiveness across performances.
\Cref{tab:evaluator_chars} reports the resulting thresholds and discriminative performance.

Importantly, the relative ordering of methods is stable across evaluators, with Kendall's $\tau = 0.949 \pm 0.048$ (all $p < 0.003$), as in \Cref{tab:evaluator_flip}.
Every evaluator recovers the same ranking, while our proposed method remains consistently at the top.
It should be noted that HVAE were trained with an additional counterfactual finetuning step, which use effectiveness as a guiding signal for optimization.
Without the additional counterfactual finetuning, HVAE's effectiveness drops significantly.
Absolute effectiveness, by contrast, varies widely on identical images: RadCF flips effusion for between $32.0\%$ and $47.2\%$ of cases depending only on which evaluator is asked, a $15$-point spread that broadly tracks evaluator AUC.
The comparative conclusions drawn from the main tables are therefore robust to the choice of evaluator, whereas any single absolute effectiveness number should be read as evaluator-relative.
We note one limitation: six evaluators are available only for effusion, and view, sex and age are re-scored with the RAD-DINO probe alone, so rank stability is established for effusion and assumed rather than verified for the remaining attributes.

\begin{table}[h]
\centering
\small
\caption{
The quality of effusion evaluators varies.
Effusion evaluators: Youden-optimal operating point and discriminative performance on the NIH-14 test split.
}
\label{tab:evaluator_chars}
\begin{tabular}{lcccc}
\toprule
Evaluator & Threshold & AUC $\uparrow$ & Sensitivity & Specificity \\
\midrule
TorchXRayVision (all)       & $0.317$ & $0.852$ & $0.867$ & $0.686$ \\
TorchXRayVision (CheXpert)  & $0.757$ & $0.756$ & $0.682$ & $0.697$ \\
TorchXRayVision (NIH)       & $0.507$ & $0.823$ & $0.793$ & $0.706$ \\
TorchXRayVision (ResNet-50) & $0.103$ & $0.864$ & $0.803$ & $0.776$ \\
CheXzero~\citep{tiu2022expert} & $0.496$ & $0.864$ & $0.878$ & $0.702$ \\
RAD-DINO linear probe       & $0.101$ & $0.853$ & $0.806$ & $0.761$ \\
\bottomrule
\end{tabular}
\end{table}

\begin{table}[h]
\centering
\small
\setlength{\tabcolsep}{4pt}
\caption{\textbf{Method rankings are consistent under all six evaluators.}
Values are the effusion flip rate on NIH-14 under each evaluator, thresholded at its Youden-optimal operating point (\Cref{tab:evaluator_chars}).
Rankings agree at Kendall's $\tau = 0.949 \pm 0.048$ (all $p < 0.003$), while absolute rates differ by as much as $0.177$ on identical images.
TXV abbreviates TorchXRayVision.
Best per column in bold.}
\label{tab:evaluator_flip}
\resizebox{\textwidth}{!}{%
\begin{tabular}{lcccccc}
\toprule
Method & CheXzero & RAD-DINO & TXV (all) & TXV (CheX) & TXV (NIH) & TXV (R50) \\
\midrule
HVAE~\citep{ribeiro2023high} (40 steps) & $0.009$ & $0.012$ & $0.010$ & $0.015$ & $0.019$ & $0.016$ \\
Diffusion~\citep{rasal2025diffusion} & $0.116$ & $0.102$ & $0.101$ & $0.105$ & $0.123$ & $0.137$ \\
\spec{s-spec}                       & $0.198$ & $0.178$ & $0.255$ & $0.227$ & $0.241$ & $0.253$ \\
Image-space flow~\citep{ribeiro2025counterfactual} & $0.266$ & $0.228$ & $0.273$ & $0.226$ & $0.235$ & $0.302$ \\
HVAE~\citep{ribeiro2023high} (20 steps) & $0.438$ & $0.261$ & $0.342$ & $0.284$ & $0.288$ & $0.303$ \\
RadCF                               & $0.454$ & $0.472$ & $0.450$ & $0.320$ & $0.416$ & $0.472$ \\
HVAE~\citep{ribeiro2023high} + counterfactual finetuning & $\mathbf{0.627}$ & $\mathbf{0.691}$ & $\mathbf{0.673}$ & $\mathbf{0.666}$ & $\mathbf{0.696}$ & $\mathbf{0.766}$ \\
\bottomrule
\end{tabular}%
}
\end{table}

\subsection{LPIPS.}\label{appendix:lpips}
We use the \texttt{lpips} package released alongside \citet{zhang2018lpips}, which computes the $\ell_2$ distance between unit-normalised intermediate activations.
We use the linearly calibrated AlexNet backbone.

\subsection{KID.}\label{appendix:kidFid}
Diverging from \citet{melistas2025benchmarking}, we use the target-domain training set rather than test-split factuals as the reference distribution for KID.
This ensures realism reflects proximity to the data manifold independently of the evaluation population.
KID is the squared MMD between Inception-v3 representations under a degree-3 polynomial kernel, computed with subset size 100 and reported as mean (std) across subsets.

\subsection{Uncertainty reporting}
\label{appendix:uncertainty}

\noindent\textbf{Binary effectiveness (flip rate).}
For discrete interventions (view, effusion, sex), effectiveness is the fraction of test images whose target attribute is successfully flipped.
We report 95\% Wald binomial confidence intervals:
$\mathrm{CI} = z\,\sqrt{p(1-p)/n}$, where $p$ is the observed flip rate, $n$ is the number of test images, and $z = 1.96$.
Effusion is reported as a view-balanced mean of the PA and AP flip rates, $p = \tfrac{1}{2}(p_{\mathrm{PA}} + p_{\mathrm{AP}})$, so that neither view dominates the score.
The two subgroups partition the test set, so their estimates are independent and
\begin{equation*}
    \mathrm{CI} = \tfrac{1}{2}\sqrt{\mathrm{CI}_{\mathrm{PA}}^{2} + \mathrm{CI}_{\mathrm{AP}}^{2}},
    \qquad
    \mathrm{CI}_{g} = z\,\sqrt{p_g(1-p_g)/n_g},
\end{equation*}
with each subgroup interval evaluated at its own size $n_g$.
When averaging flip rates across $k$ interventions (e.g.\ in \Cref{fig:data_efficiency}), we propagate per-intervention CIs as $\mathrm{CI}_{\mathrm{mean}} = \sqrt{\sum_{i=1}^{k} \mathrm{CI}_i^{2}}\,/\,k$.

\noindent\textbf{Continuous per-image metrics (LPIPS, age MAE).}
Composition LPIPS and reversibility LPIPS are computed per image; we report the median over the test set with one standard deviation of the per-image values in parentheses.
Age MAE is computed per image and averaged (mean) over the test set, reported with one standard deviation in parentheses.
When averaging across $k$ interventions, we propagate standard deviations as $\sigma_{\mathrm{mean}} = \sqrt{\sum_{i=1}^{k} \sigma_i^{2}}\,/\,k$, assuming independence across intervention families.

\noindent\textbf{KID.}
Kernel Inception Distance is estimated from random sub-samples of real and generated feature vectors.
We report one standard deviation over sub-samples in parentheses, as returned by the \texttt{torchmetrics} KID implementation.
Cross-intervention KID in \Cref{fig:data_efficiency} propagates per-intervention standard deviations using the same formula as above.

\section{Data-efficiency tables}
\label{appendix:data_efficiency_tables}

\begin{table*}[t]
\centering
\caption{\textbf{Comprehensive data efficiency results across all interventions.}
Effectiveness is the classifier flip rate for view, effusion, and sex ($\uparrow$) and judge MAE for age ($\downarrow$).
Reversibility (Rev.) is round-trip LPIPS ($\downarrow$). Composition (Comp.) is null-intervention LPIPS ($\downarrow$).
Realism is KID $\times 10^{2}$ ($\downarrow$).
Binary rates report 95\% binomial CI; LPIPS and age MAE report std in parentheses.}
\label{tab:data_efficiency_full}
\small
\setlength{\tabcolsep}{4pt}
\resizebox{\textwidth}{!}{%
\begin{tabular}{ll cccc cccc c c}
\toprule
 & & \multicolumn{4}{c}{Effectiveness} & \multicolumn{4}{c}{Reversibility (LPIPS\,$\downarrow$)} & Comp. & Realism \\
\cmidrule(lr){3-6} \cmidrule(lr){7-10} \cmidrule(lr){11-11} \cmidrule(lr){12-12}
Method & Data & $\doo(v)\!\uparrow$ & $\doo(e)\!\uparrow$ & $\doo(s)\!\uparrow$ & $\doo(a)\!\downarrow$ & $\doo(v)$ & $\doo(e)$ & $\doo(s)$ & $\doo(a)$ & LPIPS\,$\downarrow$ & KID\,$\downarrow$ \\
\midrule
 \multirow{10}{*}{\shortstack[l]{RadCF\\\spec{sa-spec}}} & 0.1\% & 0.561$_{\,(0.031)}$ & 0.350$_{\,(0.030)}$ & 0.340$_{\,(0.029)}$ & 0.223$_{\,(0.137)}$ & 0.090$_{\,(0.030)}$ & 0.066$_{\,(0.020)}$ & 0.091$_{\,(0.030)}$ & 0.053$_{\,(0.019)}$ & 0.025$_{\,(0.008)}$ & 4.46$_{\,(0.28)}$ \\
  & 0.2\% & 0.561$_{\,(0.031)}$ & 0.315$_{\,(0.030)}$ & 0.230$_{\,(0.026)}$ & 0.231$_{\,(0.136)}$ & 0.069$_{\,(0.021)}$ & 0.060$_{\,(0.017)}$ & 0.068$_{\,(0.025)}$ & 0.048$_{\,(0.015)}$ & 0.026$_{\,(0.008)}$ & 3.97$_{\,(0.28)}$ \\
  & 0.5\% & 0.812$_{\,(0.024)}$ & 0.548$_{\,(0.032)}$ & 0.440$_{\,(0.031)}$ & 0.229$_{\,(0.137)}$ & 0.077$_{\,(0.039)}$ & 0.057$_{\,(0.031)}$ & 0.070$_{\,(0.038)}$ & 0.042$_{\,(0.018)}$ & 0.023$_{\,(0.007)}$ & 4.52$_{\,(0.27)}$ \\
  & 1\% & 0.922$_{\,(0.017)}$ & 0.558$_{\,(0.032)}$ & 0.676$_{\,(0.029)}$ & 0.197$_{\,(0.124)}$ & 0.087$_{\,(0.057)}$ & 0.066$_{\,(0.052)}$ & 0.077$_{\,(0.054)}$ & 0.044$_{\,(0.034)}$ & 0.021$_{\,(0.007)}$ & 3.38$_{\,(0.27)}$ \\
  & 2\% & 0.932$_{\,(0.016)}$ & 0.657$_{\,(0.030)}$ & 0.795$_{\,(0.025)}$ & 0.188$_{\,(0.129)}$ & 0.091$_{\,(0.069)}$ & 0.047$_{\,(0.056)}$ & 0.064$_{\,(0.066)}$ & 0.040$_{\,(0.053)}$ & 0.016$_{\,(0.020)}$ & 3.82$_{\,(0.28)}$ \\
  & 5\% & 0.993$_{\,(0.005)}$ & 0.574$_{\,(0.032)}$ & 0.913$_{\,(0.017)}$ & 0.132$_{\,(0.090)}$ & 0.077$_{\,(0.072)}$ & 0.040$_{\,(0.049)}$ & 0.053$_{\,(0.062)}$ & 0.038$_{\,(0.055)}$ & 0.016$_{\,(0.019)}$ & 1.39$_{\,(0.20)}$ \\
  & 10\% & 0.992$_{\,(0.006)}$ & 0.429$_{\,(0.032)}$ & 0.805$_{\,(0.025)}$ & 0.152$_{\,(0.096)}$ & 0.068$_{\,(0.062)}$ & 0.026$_{\,(0.030)}$ & 0.034$_{\,(0.046)}$ & 0.026$_{\,(0.051)}$ & 0.015$_{\,(0.009)}$ & 0.73$_{\,(0.14)}$ \\
  & 25\% & 0.986$_{\,(0.007)}$ & 0.405$_{\,(0.031)}$ & 0.793$_{\,(0.025)}$ & 0.155$_{\,(0.098)}$ & 0.059$_{\,(0.060)}$ & 0.024$_{\,(0.025)}$ & 0.032$_{\,(0.049)}$ & 0.026$_{\,(0.046)}$ & 0.015$_{\,(0.007)}$ & 0.69$_{\,(0.13)}$ \\
  & 50\% & 0.957$_{\,(0.013)}$ & 0.435$_{\,(0.032)}$ & 0.723$_{\,(0.028)}$ & 0.175$_{\,(0.120)}$ & 0.039$_{\,(0.035)}$ & 0.021$_{\,(0.008)}$ & 0.024$_{\,(0.017)}$ & 0.023$_{\,(0.021)}$ & 0.014$_{\,(0.004)}$ & 0.72$_{\,(0.13)}$ \\
  & 100\% & 0.995$_{\,(0.004)}$ & 0.561$_{\,(0.032)}$ & 0.867$_{\,(0.021)}$ & 0.133$_{\,(0.085)}$ & 0.051$_{\,(0.058)}$ & 0.023$_{\,(0.029)}$ & 0.030$_{\,(0.044)}$ & 0.024$_{\,(0.042)}$ & 0.014$_{\,(0.010)}$ & 0.64$_{\,(0.12)}$ \\
\midrule
 \multirow{10}{*}{RadCF} & 0.1\% & 0.621$_{\,(0.030)}$ & 0.228$_{\,(0.027)}$ & 0.327$_{\,(0.029)}$ & 0.241$_{\,(0.140)}$ & 0.103$_{\,(0.023)}$ & 0.055$_{\,(0.010)}$ & 0.112$_{\,(0.028)}$ & 0.053$_{\,(0.009)}$ & 0.034$_{\,(0.006)}$ & 13.30$_{\,(0.37)}$ \\
  & 0.2\% & 0.637$_{\,(0.030)}$ & 0.308$_{\,(0.030)}$ & 0.437$_{\,(0.031)}$ & 0.241$_{\,(0.139)}$ & 0.141$_{\,(0.065)}$ & 0.082$_{\,(0.041)}$ & 0.134$_{\,(0.071)}$ & 0.053$_{\,(0.010)}$ & 0.034$_{\,(0.007)}$ & 13.43$_{\,(0.40)}$ \\
  & 0.5\% & 0.741$_{\,(0.027)}$ & 0.497$_{\,(0.032)}$ & 0.564$_{\,(0.031)}$ & 0.239$_{\,(0.136)}$ & 0.102$_{\,(0.043)}$ & 0.070$_{\,(0.031)}$ & 0.101$_{\,(0.057)}$ & 0.042$_{\,(0.009)}$ & 0.027$_{\,(0.006)}$ & 11.57$_{\,(0.40)}$ \\
  & 1\% & 0.848$_{\,(0.022)}$ & 0.523$_{\,(0.032)}$ & 0.624$_{\,(0.030)}$ & 0.239$_{\,(0.137)}$ & 0.085$_{\,(0.047)}$ & 0.056$_{\,(0.033)}$ & 0.076$_{\,(0.051)}$ & 0.039$_{\,(0.018)}$ & 0.024$_{\,(0.011)}$ & 5.51$_{\,(0.32)}$ \\
  & 2\% & 0.810$_{\,(0.024)}$ & 0.521$_{\,(0.032)}$ & 0.661$_{\,(0.029)}$ & 0.208$_{\,(0.124)}$ & 0.050$_{\,(0.036)}$ & 0.039$_{\,(0.030)}$ & 0.047$_{\,(0.039)}$ & 0.029$_{\,(0.025)}$ & 0.018$_{\,(0.009)}$ & 3.03$_{\,(0.24)}$ \\
  & 5\% & 0.684$_{\,(0.029)}$ & 0.523$_{\,(0.032)}$ & 0.562$_{\,(0.031)}$ & 0.173$_{\,(0.113)}$ & 0.036$_{\,(0.030)}$ & 0.032$_{\,(0.025)}$ & 0.033$_{\,(0.024)}$ & 0.028$_{\,(0.023)}$ & 0.017$_{\,(0.011)}$ & 2.17$_{\,(0.21)}$ \\
  & 10\% & 0.964$_{\,(0.012)}$ & 0.735$_{\,(0.028)}$ & 0.868$_{\,(0.021)}$ & 0.120$_{\,(0.088)}$ & 0.112$_{\,(0.089)}$ & 0.057$_{\,(0.078)}$ & 0.077$_{\,(0.081)}$ & 0.056$_{\,(0.088)}$ & 0.016$_{\,(0.021)}$ & 2.34$_{\,(0.26)}$ \\
  & 25\% & 0.993$_{\,(0.005)}$ & 0.592$_{\,(0.031)}$ & 0.940$_{\,(0.015)}$ & 0.095$_{\,(0.068)}$ & 0.099$_{\,(0.073)}$ & 0.039$_{\,(0.049)}$ & 0.067$_{\,(0.063)}$ & 0.053$_{\,(0.072)}$ & 0.017$_{\,(0.022)}$ & 0.51$_{\,(0.13)}$ \\
  & 50\% & 0.992$_{\,(0.006)}$ & 0.573$_{\,(0.032)}$ & 0.936$_{\,(0.015)}$ & 0.096$_{\,(0.069)}$ & 0.109$_{\,(0.073)}$ & 0.034$_{\,(0.048)}$ & 0.056$_{\,(0.062)}$ & 0.042$_{\,(0.077)}$ & 0.016$_{\,(0.025)}$ & 0.48$_{\,(0.11)}$ \\
  & 100\% & 0.998$_{\,(0.003)}$ & 0.633$_{\,(0.031)}$ & 0.952$_{\,(0.013)}$ & 0.087$_{\,(0.066)}$ & 0.104$_{\,(0.081)}$ & 0.034$_{\,(0.051)}$ & 0.061$_{\,(0.068)}$ & 0.039$_{\,(0.083)}$ & 0.015$_{\,(0.025)}$ & 0.40$_{\,(0.10)}$ \\
\bottomrule
\end{tabular}
}
\end{table*}

\clearpage

\section{Ablation studies}
\label{appendix:ablations}

\subsection{LoRA rank}
\label{appendix:rank_ablation}

\begin{figure}[t]
  \centering
  \includegraphics[width=\linewidth]{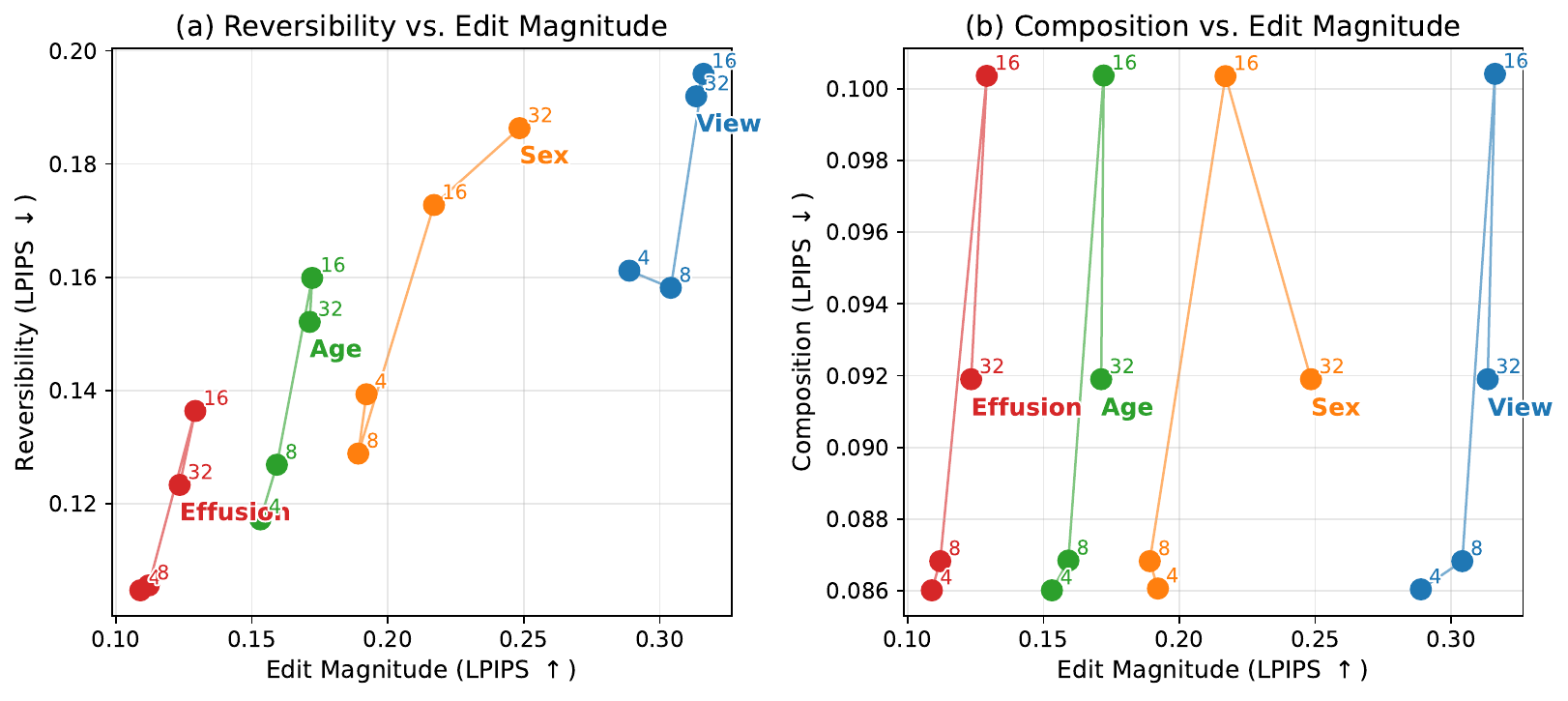}
  \caption{%
    LoRA rank ablation ($r \in \{4, 8, 16, 32\}$, frozen VAE, 100\% NIH14).
    Each panel shows counterfactual soundness metrics as a function of
    edit magnitude (CF--Orig LPIPS, higher $=$ larger edit).
    \textbf{(a)} Reversibility vs.\ edit magnitude.
    \textbf{(b)} Composition vs.\ edit magnitude.
    Points of the same colour correspond to the same intervention;
    rank labels indicate $r{=}4,8,16,32$ along each curve.
    Higher rank enables larger edits with better reversibility and
    composition; $r{=}32$ is used in all main experiments.
  }
  \label{fig:rank_tradeoff}
\end{figure}

\Cref{fig:rank_tradeoff} shows the effect of LoRA rank on the
trade-off between edit magnitude, reversibility, and composition
across all four interventions (frozen VAE, 100\% NIH14 training data).

\subsection{Latent-space decomposition of the \spec{sa-spec} gain}
\label{appendix:latent_drift}

\Cref{tab:vae_isolation} shows that adapting the autoencoder improves composition and reversibility under acquisition shift, but image-space LPIPS conflates three error sources: autoencoder reconstruction, vector-field error, and Euler integration error.
We separate them by measuring the autoencoder round trip $\mathrm{LPIPS}(\x, D_{\phi_2}(E_{\phi_1}(\x)))$ and the latent ODE round-trip drift at two solver resolutions (\Cref{tab:latent_drift}).

In domain the two variants are indistinguishable on every quantity, as expected when there is no acquisition mismatch to correct.
Under shift the autoencoder reconstruction error halves ($0.0968$ to $0.0474$, $p < 0.0001$), and this gap does not close with finer integration.
The apparent latent-drift difference behaves differently: it is present at $200$ solver steps ($p = 0.005$), but at $1000$ steps both models fall by roughly an order of magnitude and become statistically indistinguishable ($p = 0.49$).
The drift gap is therefore integration error rather than a property of the learned field, whereas the reconstruction gap persists.
This supports attributing the \spec{sa-spec} improvement to target-domain representation error rather than to transport capacity, consistent with the finding in \Cref{tab:vae_isolation} that fully fine-tuning the flow alone does not recover the gain.

\begin{table}[t]
\centering
\small
\caption{
Latent-space decomposition of the round-trip error highlights \spec{sa-spec}'s impact on reducing target-domain representation error over \spec{s-spec}.
Brackets give $95\%$ bootstrap confidence intervals; $p$-values are paired tests over patients.
$z$-drift is the latent round-trip error at $200$ and $1000$ Euler steps.
}
\label{tab:latent_drift}
\begin{tabular}{lccr}
\toprule
Metric & \spec{s-spec} & \spec{sa-spec} & $p$ \\
\midrule
\multicolumn{4}{l}{\textit{In-domain} (CheXpert\,$\rightarrow$\,CheXpert)} \\
\addlinespace[1pt]
VAE LPIPS $\downarrow$       & $0.0734$ {\scriptsize$[.0730,.0740]$}       & $0.0732$ {\scriptsize$[.0730,.0740]$}       & $0.681$ \\
$z$-drift @200 $\downarrow$  & $0.00065$ {\scriptsize$[.00047,.00099]$}    & $0.00060$ {\scriptsize$[.00046,.00082]$}    & $0.776$ \\
$z$-drift @1000 $\downarrow$ & $0.000025$ {\scriptsize$[.000018,.000037]$} & $0.000020$ {\scriptsize$[.000017,.000025]$} & $0.700$ \\
\midrule
\multicolumn{4}{l}{\textit{Acquisition shift} (CheXpert\,$\rightarrow$\,NIH-14)} \\
\addlinespace[1pt]
VAE LPIPS $\downarrow$       & $0.0968$ {\scriptsize$[.0920,.1020]$}   & $\mathbf{0.0474}$ {\scriptsize$[.0470,.0480]$}  & $<\!0.0001$ \\
$z$-drift @200 $\downarrow$  & $0.00622$ {\scriptsize$[.0040,.0090]$}  & $\mathbf{0.00291}$ {\scriptsize$[.0020,.0040]$} & $0.005$ \\
$z$-drift @1000 $\downarrow$ & $0.00061$ {\scriptsize$[.00005,.0017]$} & $0.00031$ {\scriptsize$[.00015,.00053]$}        & $0.490$ \\
\bottomrule
\end{tabular}
\end{table}

\subsection{Adapter spectrum and rank sufficiency}
\label{appendix:lora_spectrum}

\Cref{appendix:rank_ablation} selects $r = 32$ by downstream performance.
To check that this budget is not exhausted, we take the singular value decomposition of each trained LoRA update $\Delta \theta' = BA$ and summarise the spectrum per layer.
Averaged over layers, $18.6$ of the $32$ available directions carry $95\%$ of the update energy, with singular-value decay $\sigma_1/\sigma_r$ between $8\times$ and $113\times$ and a relative update magnitude $\lVert \Delta \theta' \rVert_F / \lVert \theta \rVert_F = 0.15$.
The adapter therefore uses a strict subset of its rank budget, so multi-attribute conditioning is not limited by rank saturation at $r = 32$ for the attribute combinations we test.

\clearpage
\section{Extending $\pa$ on BRAX}
\label{appendix:brax}

\begin{table*}[t]
\centering
\caption{\textbf{\spec{sa-spec} maintains identity preservation when extending $\pa$ in a new target domain (BRAX \citep{brax}).}
Each row reports one intervention.
Effectiveness is measured by pseudo-oracles.
For the intervened attribute, the corresponding pseudo-oracle metric measures target success. For non-intervened attributes, the metric measures preservation of the original.
KID is scaled by $\times 10^{2}$.}
\label{tab:brax_results}
\small
\setlength{\tabcolsep}{4pt}
\resizebox{\textwidth}{!}{%
\begin{tabular}{l ccccc c c c}
\toprule
 & \multicolumn{5}{c}{Effectiveness} & Composition & Reversibility & Realism \\
\cmidrule(lr){2-6}
\cmidrule(lr){7-7}
\cmidrule(lr){8-8}
\cmidrule(lr){9-9}
Interv. & $\text{Acc}(v)\uparrow$ & $\text{Acc}(e)\uparrow$ & $\text{Acc}(s)\uparrow$ & $\text{MAE}(a)\downarrow$ & $\text{Acc}(d)\uparrow$ & LPIPS\,$\downarrow$ & LPIPS\,$\downarrow$ & KID\,$\downarrow$ \\
\midrule
 $\doo(v)$  & 0.870$_{\,(0.012)}$ & 0.613$_{\,(0.018)}$ & 0.785$_{\,(0.015)}$ & 0.094$_{\,(0.078)}$ & 0.668$_{\,(0.017)}$ & 0.018$_{\,(0.021)}$ & 0.019$_{\,(0.034)}$ & 2.67$_{\,(0.73)}$ \\
 $\doo(e)$  & 0.988$_{\,(0.004)}$ & 0.509$_{\,(0.013)}$ & 0.913$_{\,(0.010)}$ & 0.061$_{\,(0.054)}$ & 0.786$_{\,(0.015)}$ & 0.018$_{\,(0.021)}$ & 0.020$_{\,(0.044)}$ & 1.14$_{\,(0.35)}$ \\
 $\doo(s)$  & 0.989$_{\,(0.004)}$ & 0.787$_{\,(0.015)}$ & 0.867$_{\,(0.012)}$ & 0.062$_{\,(0.056)}$ & 0.732$_{\,(0.016)}$ & 0.018$_{\,(0.021)}$ & 0.019$_{\,(0.040)}$ & 0.49$_{\,(0.24)}$ \\
 $\doo(a)$  & 0.974$_{\,(0.006)}$ & 0.761$_{\,(0.015)}$ & 0.907$_{\,(0.011)}$ & 0.100$_{\,(0.081)}$ & 0.776$_{\,(0.015)}$ & 0.018$_{\,(0.021)}$ & 0.019$_{\,(0.036)}$ & 0.75$_{\,(0.23)}$ \\
 $\doo(d)$  & 0.985$_{\,(0.004)}$ & 0.774$_{\,(0.015)}$ & 0.899$_{\,(0.011)}$ & 0.074$_{\,(0.070)}$ & 0.729$_{\,(0.016)}$ & 0.018$_{\,(0.021)}$ & 0.019$_{\,(0.042)}$ & 0.78$_{\,(0.25)}$ \\
\bottomrule
\end{tabular}
}
\end{table*}

\clearpage
\section{Downstream utility experiments}
\label{appendix:utilityExperiments}

\noindent\textbf{Biased dataset construction.}
We construct biased training and test sets from NIH-14 by subsampling from the patient pool.
Each bias level sets the probability that an effusion-positive case is assigned the PA view, so the effusion rate per view reflects how strongly view predicts disease (\Cref{tab:utility_bias}).
The audit experiment spans ten levels of spurious view-effusion correlation from 0.50 to 0.95; the debiasing experiment uses three representative levels (0.50, 0.75, 0.95).

\begin{table}[t]
\centering
\small
\caption{\textbf{Training dataset bias construction.}
Each level is subsampled from the same patient pool to create a different level of spurious view-effusion correlation.
$P(e{=}1)$ is the effusion rate per view, and a larger gap between views means view predicts effusion more strongly.}
\label{tab:utility_bias}
\setlength{\tabcolsep}{5pt}
\begin{tabular}{llrrr}
\toprule
Bias & $v$ & $e=0$ & $e=1$ & $P(e{=}1)$ \\
\midrule
\multirow{2}{*}{0.50 ($N=4{,}945$)} & AP & 2{,}671 & 294 & 9.9\% \\
 & PA & 1{,}626 & 354 & 17.9\% \\
\midrule
\multirow{2}{*}{0.55 ($N=5{,}061$)} & AP & 2{,}949 & 272 & 8.4\% \\
 & PA & 1{,}450 & 390 & 21.2\% \\
\midrule
\multirow{2}{*}{0.60 ($N=5{,}168$)} & AP & 3{,}213 & 239 & 6.9\% \\
 & PA & 1{,}290 & 426 & 24.8\% \\
\midrule
\multirow{2}{*}{0.65 ($N=5{,}265$)} & AP & 3{,}468 & 215 & 5.8\% \\
 & PA & 1{,}126 & 456 & 28.8\% \\
\midrule
\multirow{2}{*}{0.70 ($N=5{,}401$)} & AP & 3{,}739 & 181 & 4.6\% \\
 & PA & 986 & 495 & 33.4\% \\
\midrule
\multirow{2}{*}{0.75 ($N=5{,}501$)} & AP & 4{,}004 & 156 & 3.8\% \\
 & PA & 808 & 533 & 39.7\% \\
\midrule
\multirow{2}{*}{0.80 ($N=5{,}580$)} & AP & 4{,}250 & 121 & 2.8\% \\
 & PA & 644 & 565 & 46.7\% \\
\midrule
\multirow{2}{*}{0.85 ($N=5{,}704$)} & AP & 4{,}541 & 88 & 1.9\% \\
 & PA & 475 & 600 & 55.8\% \\
\midrule
\multirow{2}{*}{0.90 ($N=5{,}812$)} & AP & 4{,}796 & 57 & 1.2\% \\
 & PA & 326 & 633 & 66.0\% \\
\midrule
\multirow{2}{*}{0.95 ($N=5{,}904$)} & AP & 5{,}053 & 31 & 0.6\% \\
 & PA & 156 & 664 & 81.0\% \\
\bottomrule
\end{tabular}
\end{table}

\noindent\textbf{Shortcut learning audit experiment.}
At each bias level, we train a DenseNet-121 classifier to predict effusion with five-fold patient-level cross validation, reporting all results as the mean and standard deviation across folds.
To probe each classifier, we generate view counterfactuals $\doo(v)$ for every test image with \spec{sa-spec} RadCF (CheXpert$\to$NIH-14).
As a random control, we additionally re-score each original under label-preserving, non-generative augmentations (horizontal flip, small affine, colour jitter), which perturb the image without altering the view.
We then compare the classifier's predictions across the original, counterfactual, and control images.

\noindent\textbf{Counterfactual augmentation experiment.}
To test whether counterfactual augmentation can reduce shortcut learning during training, we compare a baseline classifier against three augmentation strategies: random upsampling, conventional random augmentation, and counterfactual augmentation.
Each strategy is size-matched to twice the biased set for a fair comparison.
Augmentation and auditing deliberately use different RadCF generation mechanisms: the view-flipped counterfactuals added to training come from \spec{s-spec} trained on NIH-14, while the auditing is carried out by the independent \spec{sa-spec} auditor.
This avoids circularity: were the same generator used for both, a classifier could appear robust simply by learning to agree with that generator's edits.
Auditing with an independent generator ensures that reduced reliance reflects robustness to the view shortcut rather than consistency with the augmentation generator.

\begin{table}[h]
\centering
\small
\caption{\textbf{Augmenting training with counterfactuals reduces view-shortcut reliance without sacrificing discrimination.}
Effusion classifiers are trained under three bias levels with each mitigation strategy (size-matched to 2N samples) and evaluated on original and view-counterfactual images.
Prediction change rate (decision threshold=0.5) measures reliance under \(\doo(v)\), and AUC$_{\text{orig}}$/AUC$_{\text{cf}}$ measure discrimination on real and counterfactual images.
View gap is the stratified subgroup difference $|\text{AUC}_{\text{AP}} - \text{AUC}_{\text{PA}}|$ on originals.
The mean and standard deviation of five-fold cross validation are reported.
Within each bias level, the best value per column is in bold.}
\label{tab:utility_debiasing}
\setlength{\tabcolsep}{5pt}
\begin{tabular}{ll rrrr}
\toprule
Bias & Mitigation & AUC$_{\text{orig}}\uparrow$ & AUC$_{\text{cf}}\uparrow$ & Change rate\,$\downarrow$ & View gap\,$\downarrow$ \\
\midrule
\multirow{4}{*}{0.50}
 & Baseline & 0.841$_{\,(0.018)}$ & 0.748$_{\,(0.021)}$ & \textbf{0.070}$_{\,(0.022)}$ & 0.081$_{\,(0.012)}$ \\
 & Random upsample & 0.831$_{\,(0.025)}$ & 0.721$_{\,(0.021)}$ & 0.214$_{\,(0.055)}$ & 0.078$_{\,(0.012)}$ \\
 & Random aug. & 0.848$_{\,(0.014)}$ & 0.748$_{\,(0.022)}$ & 0.088$_{\,(0.017)}$ & \textbf{0.076}$_{\,(0.013)}$ \\
 & Counterf. aug. & \textbf{0.852}$_{\,(0.009)}$ & \textbf{0.805}$_{\,(0.007)}$ & 0.085$_{\,(0.024)}$ & 0.081$_{\,(0.006)}$ \\
\midrule
\multirow{4}{*}{0.75}
 & Baseline & 0.816$_{\,(0.014)}$ & 0.663$_{\,(0.021)}$ & 0.253$_{\,(0.042)}$ & \textbf{0.062}$_{\,(0.023)}$ \\
 & Random upsample & \textbf{0.821}$_{\,(0.013)}$ & 0.703$_{\,(0.024)}$ & 0.307$_{\,(0.079)}$ & 0.078$_{\,(0.029)}$ \\
 & Random aug. & 0.820$_{\,(0.015)}$ & 0.645$_{\,(0.041)}$ & 0.263$_{\,(0.079)}$ & 0.077$_{\,(0.022)}$ \\
 & Counterf. aug. & 0.819$_{\,(0.017)}$ & \textbf{0.760}$_{\,(0.011)}$ & \textbf{0.101}$_{\,(0.014)}$ & 0.078$_{\,(0.034)}$ \\
\midrule
\multirow{4}{*}{0.95}
 & Baseline & 0.703$_{\,(0.036)}$ & 0.531$_{\,(0.039)}$ & 0.593$_{\,(0.197)}$ & 0.084$_{\,(0.054)}$ \\
 & Random upsample & \textbf{0.734}$_{\,(0.024)}$ & 0.581$_{\,(0.024)}$ & 0.468$_{\,(0.092)}$ & \textbf{0.030}$_{\,(0.012)}$ \\
 & Random aug. & 0.714$_{\,(0.021)}$ & 0.528$_{\,(0.034)}$ & 0.655$_{\,(0.062)}$ & 0.036$_{\,(0.027)}$ \\
 & Counterf. aug. & 0.712$_{\,(0.042)}$ & \textbf{0.683}$_{\,(0.027)}$ & \textbf{0.115}$_{\,(0.020)}$ & 0.071$_{\,(0.032)}$ \\
\bottomrule
\end{tabular}
\end{table}

\begin{figure*}[t]
\centering
\begin{subfigure}[t]{0.32\textwidth}
\centering
\includegraphics[width=\textwidth]{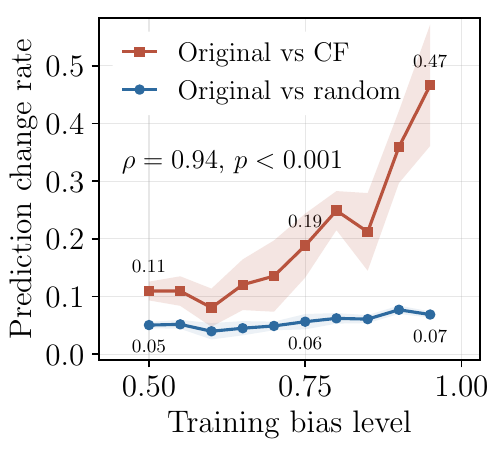}
\caption{Prediction change rate.}
\label{fig:shortcut_flips}
\end{subfigure}
\quad\quad
\begin{subfigure}[t]{0.32\textwidth}
\centering
\includegraphics[width=\textwidth]{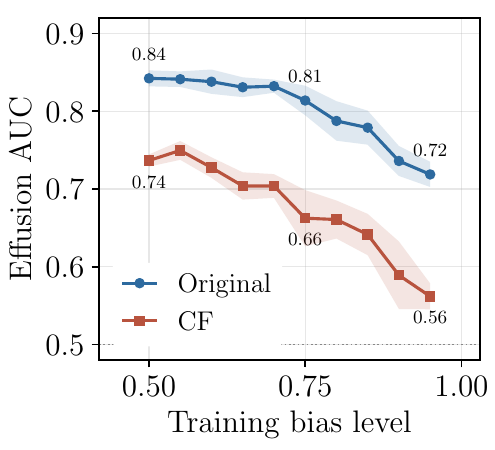}
\caption{Prediction AUC.}
\label{fig:shortcut_auc}
\end{subfigure}
\caption{\textbf{Paired original-counterfactual probing exposes shortcut learning that accuracy metrics miss.}
An effusion classifier is trained at ten view-effusion bias levels and probed with counterfactuals $\doo(v)$ and a random, non-generative control.
\textbf{(a)}~The change rate rises with the imposed bias while a random control (non-generative augmentation of the original) stays flat, indicating the signal reflects the view intervention rather than generic sensitivity to image perturbation.
\textbf{(b)}~Counterfactual AUC falls toward chance while original AUC declines more gradually, indicating that biased models increasingly rely on view-correlated features.}
\label{fig:shortcut_audit}
\end{figure*}

\clearpage
\section{Discussion on the break of Markovian assumption}
\label{appendix:probe}

The mechanism in \Cref{eq:scm_mechanism} assumes $\uu \perp \pa$, a standard assumption in deep structural causal modelling \citep{pearl2009causality,pawlowski2020deepscm,monteiro2023axiomatic,ribeiro2023high,rasal2025diffusion,ribeiro2025counterfactual}.
Two questions it raises are worth separating: whether our estimator induces a violation, and whether the data satisfy the assumption in the first place.

\noindent\textbf{The model does not induce a violation.}
A batch-wise optimal-transport coupling \citep{tong2024improving}, for instance, pairs noise with data within each minibatch and thereby ties the inferred terminal noise to the sampled batch, which can reintroduce dependence between the abducted noise and the parents.
RadCF interpolates from a parent-independent Gaussian prior under an independent coupling: in \Cref{eq:fm_loss} the terminal noise is drawn as $\z_1 \sim \mathcal{N}(0, \identity)$ independently of both the data encoding $\z_0$ and the conditioning $\cc = \pa$, so prior and training coupling are parent-independent by construction.

\noindent\textbf{We cannot rule out confounding in the data.}
Causal sufficiency is not testable from i.i.d.\ images and labels \citep{pan2024counterfactual}.
It underpins essentially all deep-SCM counterfactual image generation \citep{pawlowski2020deepscm,monteiro2023axiomatic,ribeiro2023high,rasal2025diffusion,ribeiro2025counterfactual}, and as in that literature we target counterfactuals relative to the specified mechanism rather than uniquely identified real-world counterfactuals.
Under unobserved confounding RadCF remains a counterfactual estimator with respect to the fitted mechanism, but is not guaranteed to recover the true clinical counterfactual.

\noindent\textbf{Measuring the residual dependence.}
To measure the extent to which the abducted noise $\uu$ retains parent information, we fit linear probes to predict each parent from $\uu$ and report held-out probe AUC (\Cref{tab:probe}).
Independence by construction during training does not guarantee independence of the \emph{abducted} noise at inference time, so we measure it directly.
Here lower is better, and $0.5$ would indicate no linearly decodable parent information.

The unconditional mechanism serves as a reference point: it has no parent conditioning, so parent-related variation has nowhere to go except $\uu$, and its probe AUCs are correspondingly high.
Specialisation lowers the predictability of every parent relative to it, indicating that conditioning moves parent-related variation out of the exogenous latent.
We still acknowledge that the abducted latent is not perfectly independent of the parents, and that the Markovian assumption may be violated to some degree in practice.
Thus, we rely on the axiomatic criteria to measure the quality of the counterfactuals relative to the fitted mechanism.

\begin{table}[h]
\centering
\small
\caption{\textbf{Conditioning moves parent information out of the exogenous latent, but not to chance.}
Held-out linear-probe AUC for predicting each parent from the abducted noise $\uu$; lower is closer to $\uu \perp \pa$, $0.5$ is chance.
The unconditional mechanism is a reference point, not an upper bound.
Best per column in bold.}
\label{tab:probe}
\begin{tabular}{lccc}
\toprule
Mechanism & View & Sex & Effusion \\
\midrule
Unconditional                                      & $0.995$ & $0.817$ & $0.636$ \\
Image-space flow~\citep{ribeiro2025counterfactual}  & $0.972$ & $0.832$ & $0.833$ \\
\spec{s-spec}                                      & $\mathbf{0.967}$ & $\mathbf{0.761}$ & $\mathbf{0.571}$ \\
\bottomrule
\end{tabular}
\end{table}

\clearpage
\section{SCM-guided propagation implementation details}
\label{appendix:scm_propagation}

To illustrate composition with an upstream SCM, we manually specify a metadata SCM over $\mathbf{m}=(s,a,e,v)$:
\begin{equation}
    a \rightarrow e \rightarrow v,
    \qquad
    \{s,a,e,v\}\rightarrow \x ,
    \label{eq:metadata_scm}
\end{equation}
where $s$ and $a$ are observed root variables, $e$ denotes effusion status, and $v$ denotes acquisition view.
The leaf mechanism $\{s,a,e,v\}\rightarrow\x$ is the specialised image mechanism and is kept fixed throughout this experiment.
Here the graph decides which parents change; \Cref{appendix:joint_capacity} instead intervenes on a fixed number of parents chosen at random, isolating the capacity of the mechanism from the structure of the graph.

We fit Bernoulli mechanisms for the two non-root metadata variables \citep{oberst2019counterfactualoffpolicyevaluationgumbelmax}:
\begin{equation}
    e
    =
    \mathbb{1}\!\left[
        u_e < \hat p_e(a)
    \right],
    \qquad
    v
    =
    \mathbb{1}\!\left[
        u_v < \hat p_v(e)
    \right],
    \qquad
    u_e,u_v\sim\mathrm{Unif}(0,1).
\end{equation}
Here, $\hat p_e(a)$ approximates $P(e=1\mid a)$ and $\hat p_v(e)$ approximates $P(v=\mathrm{AP}\mid e)$.
Each mechanism is parameterised by a separate two-layer perceptron and trained with binary cross-entropy on the observational NIH-14 training split.
No image-model parameters are updated.

The edge $a\rightarrow e$ represents an assumed dependence of effusion prevalence on age.
The edge $e\rightarrow v$ represents an assumed acquisition pathway in which the presence of effusion may be associated with reduced patient mobility and therefore a greater probability of AP acquisition.
These edges are manually specified, rather than identified from the observational metadata.
Counterfactual inference follows abduction--action--prediction \citep{pearl2009causality,pawlowski2020deepscm}.

\Cref{sec:result_utility} evaluates propagation under $\doo(e)$.
In this query, the intervention replaces the effusion mechanism, severing the incoming edge $a\rightarrow e$, while allowing the intervention to propagate through $e\rightarrow v$. Although the SCM also supports propagation under $\doo(a)$, the fitted $a\rightarrow e$ mechanism produces too few realised effusion changes under our age interventions
to support stable image-level evaluation.

\begin{table*}[t]
\centering
\caption{\textbf{SCM-guided propagation realises counterfactual image generation under a specified SCM.}
Both query types are evaluated on the same randomly selected $1{,}500$ samples, stratified by whether the SCM-guided query flips the view.
Effectiveness is measured by the pseudo-oracles of \Cref{tab:main_results}.
KID is scaled by $\times 10^{2}$.
The better value per column within each block is in bold.}
\label{tab:scm_propagation}
\small
\setlength{\tabcolsep}{4pt}
\resizebox{\textwidth}{!}{%
\begin{tabular}{ll cccc c c c}
\toprule
 & & \multicolumn{4}{c}{Effectiveness} & Composition & Reversibility & Realism \\
\cmidrule(lr){3-6}
\cmidrule(lr){7-7}
\cmidrule(lr){8-8}
\cmidrule(lr){9-9}
Cases & Query & $\text{Acc}(v)\uparrow$ & $\text{Acc}(e)\uparrow$ & $\text{Acc}(s)\uparrow$ & $\text{MAE}(a)\downarrow$ & LPIPS\,$\downarrow$ & LPIPS\,$\downarrow$ & KID\,$\downarrow$ \\
\midrule
 \multirow{2}{*}{\shortstack[l]{View changed\\$n=307$}}
 & Direct & 1.000$_{\,(0.000)}$ & 0.827$_{\,(0.042)}$ & 0.961$_{\,(0.022)}$ & \textbf{0.042}$_{\,(0.051)}$ & 0.014$_{\,(0.009)}$ & \textbf{0.021}$_{\,(0.025)}$ & \textbf{0.30}$_{\,(0.20)}$ \\

 & SCM-guided & 1.000$_{\,(0.000)}$ & 0.827$_{\,(0.042)}$ & \textbf{0.964}$_{\,(0.021)}$ & 0.067$_{\,(0.068)}$ & 0.014$_{\,(0.009)}$ & 0.038$_{\,(0.050)}$ & 4.80$_{\,(0.70)}$ \\
\midrule
 \multirow{2}{*}{\shortstack[l]{View unchanged\\$n=1{,}193$}}
 & Direct & 0.995$_{\,(0.004)}$ & 0.873$_{\,(0.019)}$ & 0.966$_{\,(0.010)}$ & 0.041$_{\,(0.045)}$ & 0.014$_{\,(0.012)}$ & 0.021$_{\,(0.025)}$ & 0.40$_{\,(0.20)}$ \\

 & SCM-guided & 0.995$_{\,(0.004)}$ & 0.873$_{\,(0.019)}$ & 0.966$_{\,(0.010)}$ & 0.041$_{\,(0.045)}$ & 0.014$_{\,(0.012)}$ & 0.021$_{\,(0.025)}$ & 0.40$_{\,(0.20)}$ \\
\bottomrule
\end{tabular}
}
\end{table*}

\clearpage
\section{Multi-attribute capacity}
\label{appendix:joint_capacity}

Our main results intervene on one parent at a time.
\Cref{tab:joint_capacity} instead reports \emph{random joint} interventions, in which $k$ of the four parents are sampled uniformly per image and intervened on simultaneously. This is a capacity stress test independent of any graph.

Per-attribute effectiveness is essentially invariant to $k$: view accuracy stays at $0.998$ to $0.993$, and effusion and sex accuracy vary by under two points across $k = 2, 3, 4$ rather than degrading.
Reversibility rises from $0.055$ to $0.069$ and KID from $0.01$ to $0.03$, both monotonically, as the counterfactual departs from the factual image along more axes.

\begin{table}[h]
\centering
\small
\setlength{\tabcolsep}{4pt}
\caption{
\textbf{Random joint interventions on $k$ simultaneously edited parents.}
Effectiveness is reported per attribute over the subset of samples in which that attribute
was among the $k$ intervened on. Composition and reversibility are means across
interventions; unlike the other tables, KID is reported unscaled here. Subscripts give standard deviations
(\Cref{appendix:uncertainty}).
}
\label{tab:joint_capacity}
\begin{tabular}{lccccccc}
\toprule
 & $\text{Acc}(v)\uparrow$ & $\text{Acc}(e)\uparrow$ & $\text{Acc}(s)\uparrow$ & $\text{MAE}(a)\downarrow$ & Comp.$\,\downarrow$ & Rev.$\,\downarrow$ & KID\,$\downarrow$ \\
\midrule
$k = 2$ & $0.998_{\,(0.044)}$ & $0.852_{\,(0.355)}$ & $0.870_{\,(0.336)}$ & $0.096_{\,(0.070)}$ & $0.016_{\,(0.010)}$ & $0.055_{\,(0.050)}$ & $0.01_{\,(0.00)}$ \\
$k = 3$ & $0.993_{\,(0.082)}$ & $0.862_{\,(0.344)}$ & $0.856_{\,(0.351)}$ & $0.108_{\,(0.079)}$ & $0.016_{\,(0.010)}$ & $0.064_{\,(0.056)}$ & $0.02_{\,(0.00)}$ \\
$k = 4$ & $0.993_{\,(0.083)}$ & $0.858_{\,(0.349)}$ & $0.887_{\,(0.317)}$ & $0.109_{\,(0.082)}$ & $0.016_{\,(0.010)}$ & $0.069_{\,(0.055)}$ & $0.03_{\,(0.01)}$ \\
\bottomrule
\end{tabular}
\end{table}

\clearpage
\section{Cross-backbone comparison with Causal-Adapter}
\label{app:causal_adapter}

\Cref{tab:radedit_results} in the main text compares \spec{m-spec} against Causal-Adapter~\citep{tong2026causaladapter} under a matched backbone: every row is built on RadEdit, so the rows differ only in how causal structure is injected. \Cref{tab:causal_adapter_full} additionally reports the authors' original implementation, which uses Stable Diffusion 1.5 as its backbone.

The SD1.5 rows are not backbone-matched, so effectiveness differences confound the pretrained backbone with the adaptation method. SD1.5 Causal-Adapter attains higher effectiveness on $\doo(v)$, $\doo(s)$ and $\doo(a)$, which we attribute mainly to its natural-image pretraining: the attributes it recovers best are global image properties rather than localised pathology. Its composition ($0.113$) and reversibility ($0.197$ mean across interventions) are nonetheless worse than \spec{m-spec} ($0.061$ / $0.167$), and its realism is markedly worse (KID $3.13$ vs.\ $1.35$). Under the matched RadEdit backbone, \spec{m-spec} improves on Causal-Adapter for every target attribute and for composition, reversibility and realism; the two exceptions are non-target preservation of sex under $\doo(v)$ ($0.947$ vs.\ $0.963$) and under $\doo(a)$ ($0.962$ vs.\ $0.976$).

\begin{table*}[t]
\centering
\caption{
\textbf{Full comparison with Causal-Adapter~\citep{tong2026causaladapter}, including the authors' original SD1.5-backbone implementation.}
Extends \Cref{tab:radedit_results} with the SD1.5 Causal-Adapter rows.
SD1.5 Causal-Adapter is pretrained on a different backbone (Stable Diffusion 1.5) from every other row (RadEdit), so differences in effectiveness confound backbone and adaptation method; RadEdit Causal-Adapter is the controlled comparison.
KID is scaled by $\times 10^{2}$.
Best per column metrics are in bold. }
\label{tab:causal_adapter_full}
\small
\setlength{\tabcolsep}{4pt}
\resizebox{\textwidth}{!}{%
\begin{tabular}{ll cccc c c c}
\toprule
 & & \multicolumn{4}{c}{Effectiveness} & Composition & Reversibility & Realism \\
\cmidrule(lr){3-6}
\cmidrule(lr){7-7}
\cmidrule(lr){8-8}
\cmidrule(lr){9-9}
Intervention & Method & $\text{Acc}(v)\uparrow$ & $\text{Acc}(e)\uparrow$ & $\text{Acc}(s)\uparrow$ & $\text{MAE}(a)\downarrow$ & LPIPS\,$\downarrow$ & LPIPS\,$\downarrow$ & KID\,$\downarrow$ \\
\midrule
 \multirow{3}{*}{$\doo(v)$}
 & RadEdit Causal-Adapter & 0.563$_{\,(0.031)}$ & 0.658$_{\,(0.029)}$ & 0.963$_{\,(0.012)}$ & 0.073$_{\,(0.061)}$ & 0.094$_{\,(0.072)}$ & 0.278$_{\,(0.091)}$ & 4.64$_{\,(0.61)}$ \\

 & SD1.5 Causal-Adapter~\citep{tong2026causaladapter} & \textbf{0.939}$_{\,(0.015)}$ & \textbf{0.776}$_{\,(0.026)}$ & \textbf{0.968}$_{\,(0.011)}$ & \textbf{0.060}$_{\,(0.050)}$ & 0.113$_{\,(0.038)}$ & 0.231$_{\,(0.059)}$ & 4.71$_{\,(0.55)}$ \\

 & RadEdit \spec{m-spec} & 0.600$_{\,(0.030)}$ & 0.670$_{\,(0.029)}$ & 0.947$_{\,(0.014)}$ & 0.067$_{\,(0.056)}$ & \textbf{0.061}$_{\,(0.057)}$ & \textbf{0.202}$_{\,(0.067)}$ & \textbf{2.69}$_{\,(0.50)}$ \\
\midrule
 \multirow{5}{*}{$\doo(e)$}
 & SDXL & 0.748$_{\,(0.027)}$ & \textbf{0.776}$_{\,(0.027)}$ & 0.680$_{\,(0.029)}$ & 0.150$_{\,(0.115)}$ & \textbf{0.015}$_{\,(0.003)}$ & 0.488$_{\,(0.099)}$ & 3.54$_{\,(0.39)}$ \\

 & RadEdit & 0.995$_{\,(0.004)}$ & 0.566$_{\,(0.031)}$ & 0.972$_{\,(0.010)}$ & 0.070$_{\,(0.056)}$ & 0.027$_{\,(0.015)}$ & \textbf{0.107}$_{\,(0.050)}$ & \textbf{1.08}$_{\,(0.24)}$ \\

 & RadEdit Causal-Adapter & 0.979$_{\,(0.009)}$ & 0.531$_{\,(0.032)}$ & 0.974$_{\,(0.010)}$ & 0.065$_{\,(0.058)}$ & 0.094$_{\,(0.072)}$ & 0.223$_{\,(0.108)}$ & 1.85$_{\,(0.42)}$ \\

 & SD1.5 Causal-Adapter~\citep{tong2026causaladapter} & 0.996$_{\,(0.004)}$ & 0.493$_{\,(0.032)}$ & 0.970$_{\,(0.011)}$ & \textbf{0.048}$_{\,(0.040)}$ & 0.113$_{\,(0.038)}$ & 0.181$_{\,(0.061)}$ & 2.79$_{\,(0.39)}$ \\

 & RadEdit \spec{m-spec} & \textbf{1.000}$_{\,(0.000)}$ & 0.682$_{\,(0.030)}$ & \textbf{0.976}$_{\,(0.009)}$ & 0.053$_{\,(0.043)}$ & 0.061$_{\,(0.057)}$ & 0.174$_{\,(0.070)}$ & 1.40$_{\,(0.39)}$ \\
\midrule
 \multirow{3}{*}{$\doo(s)$}
 & RadEdit Causal-Adapter & 0.986$_{\,(0.007)}$ & 0.768$_{\,(0.026)}$ & 0.160$_{\,(0.023)}$ & 0.063$_{\,(0.054)}$ & 0.094$_{\,(0.072)}$ & 0.186$_{\,(0.106)}$ & 1.64$_{\,(0.45)}$ \\

 & SD1.5 Causal-Adapter~\citep{tong2026causaladapter} & 0.996$_{\,(0.004)}$ & \textbf{0.871}$_{\,(0.021)}$ & \textbf{0.729}$_{\,(0.028)}$ & \textbf{0.047}$_{\,(0.039)}$ & 0.113$_{\,(0.038)}$ & 0.190$_{\,(0.063)}$ & 2.46$_{\,(0.42)}$ \\

 & RadEdit \spec{m-spec} & \textbf{0.997}$_{\,(0.003)}$ & 0.839$_{\,(0.023)}$ & 0.547$_{\,(0.031)}$ & 0.048$_{\,(0.039)}$ & \textbf{0.061}$_{\,(0.057)}$ & \textbf{0.150}$_{\,(0.070)}$ & \textbf{0.75}$_{\,(0.26)}$ \\
\midrule
 \multirow{3}{*}{$\doo(a)$}
 & RadEdit Causal-Adapter & 0.981$_{\,(0.008)}$ & 0.790$_{\,(0.025)}$ & \textbf{0.976}$_{\,(0.009)}$ & 0.200$_{\,(0.133)}$ & 0.094$_{\,(0.071)}$ & 0.213$_{\,(0.107)}$ & 1.67$_{\,(0.39)}$ \\

 & SD1.5 Causal-Adapter~\citep{tong2026causaladapter} & \textbf{0.997}$_{\,(0.003)}$ & \textbf{0.874}$_{\,(0.021)}$ & 0.975$_{\,(0.010)}$ & \textbf{0.172}$_{\,(0.108)}$ & 0.113$_{\,(0.038)}$ & 0.185$_{\,(0.059)}$ & 2.57$_{\,(0.37)}$ \\

 & RadEdit \spec{m-spec} & 0.993$_{\,(0.005)}$ & 0.848$_{\,(0.022)}$ & 0.962$_{\,(0.012)}$ & 0.191$_{\,(0.126)}$ & \textbf{0.062}$_{\,(0.058)}$ & \textbf{0.143}$_{\,(0.077)}$ & \textbf{0.57}$_{\,(0.22)}$ \\
\bottomrule
\end{tabular}
}
\end{table*}

\end{document}